\documentclass[twoside]{article}
\usepackage[preprint]{aistats2027}
\usepackage{graphicx,algorithm,algorithmic,booktabs,wrapfig,varwidth}
\usepackage{amsmath,amssymb,amsthm,natbib}
\newtheorem{proposition}{Proposition}
\usepackage[hidelinks]{hyperref}
\usepackage{url,xcolor,tikz,pgfplots}
\pgfplotsset{compat=1.18}
\usetikzlibrary{arrows.meta,positioning,fit,backgrounds}
\definecolor{cbBlue}{HTML}{2F6DAB}
\definecolor{cbOrange}{HTML}{C2542B}
\definecolor{cbGreen}{HTML}{3E8E5A}
\definecolor{cbGray}{HTML}{6E6E6E}
\definecolor{cbPurple}{HTML}{7B4FA3}
\newcommand{\num}[1]{#1}
\hypersetup{
  hidelinks,
  pdftitle={EvolveNet: Collaborative Harness Evolution for Agent Self-Improvement},
  pdfauthor={Jun Nie, Yonggang Zhang, Qianshu Cai, Yiu-ming Cheung, Xinmei Tian, Bo Han}
}
\begin{document}
\runningtitle{EvolveNet: Collaborative Harness Evolution for Agent Self-Improvement}
\runningauthor{EvolveNet: Collaborative Harness Evolution for Agent Self-Improvement}
\twocolumn[
\aistatstitle{EvolveNet: Collaborative Harness Evolution for Agent Self-Improvement}
\aistatsauthor{
Jun Nie$^{1,2}$ \qquad Yonggang Zhang$^{3,*}$ \qquad Qianshu Cai$^{2,3}$ \\
\bfseries Yiu-ming Cheung$^{1}$ \qquad Xinmei Tian$^{2}$ \qquad Bo Han$^{1,*}$
}
\aistatsaddress{
$^{1}$Hong Kong Baptist University \\
$^{2}$University of Science and Technology of China \\
$^{3}$The Hong Kong University of Science and Technology \\[0.5em]
}]
\begingroup
\makeatletter
\renewcommand{\@makefntext}[1]{\noindent\raggedright\@makefnmark#1}
\makeatother
\renewcommand{\thefootnote}{\fnsymbol{footnote}}
\footnotetext[1]{Corresponding authors.\\
  Code: \url{https://github.com/junnie00/EvolveNet}}
\endgroup
\begin{abstract}
An LLM agent can acquire persistent improvements by evolving its harness---the executable program governing model calls, tool use, verification, and recovery---without updating model weights. When the same harness serves users with different workloads, isolated evolution limits the sharing of useful discoveries, while centralized adaptation can introduce conflicting task-specific changes. We introduce \emph{EvolveNet}, a framework for collaborative harness evolution that connects local specialization with shared capability accumulation. Clients evolve the same broadcast harness on their local workloads in parallel, and a server composes their adaptations into an updated shared program. Redistribution makes accepted discoveries available to all clients, allowing subsequent local evolution to build on improvements accumulated across the collaboration. EvolveNet supports this process through evidence-guided, scope-typed program aggregation and behavioral validation. Our theoretical analysis provides sufficient conditions for retaining and combining complementary improvements despite program interactions. Experiments across five settings show improved shared-harness performance, with ablation and behavioral analyses supporting the contribution of composition beyond specialist selection alone.
\end{abstract}

\section{Introduction}

An LLM agent's capabilities depend on both its model and the program around it. This \emph{harness} determines how the agent constructs context, invokes tools, checks intermediate results, and recovers from failure. Optimizing the harness therefore provides a program-level route to agent self-improvement without updating model weights. Prompt optimization, workflow search, and automated agent design explore different parts of this design space \citep{fernando2024promptbreeder,yang2024opro,hu2025adas,zhang2025dgm}; recent harness-evolution methods directly revise the executable control program using feedback from its executions \citep{he2025evotest,lee2026metaharness,zhang2026selfharness,nie2026tthe}. Such revisions can encode experience in reusable tool-use, verification, and recovery procedures, allowing useful discoveries to influence subsequent executions.

When a common harness serves multiple users, its executions on different databases, code libraries, repositories, and tool environments expose different opportunities for improvement. Independent local evolution can exploit these opportunities, but its benefits remain confined to individual clients unless their discoveries are shared. A useful repair found by one client can remain absent from another client's program even when it would apply there. More users and more execution experience therefore do not automatically produce a stronger shared agent: local evolution may generate increasingly specialized versions without accumulating their complementary strengths in the common harness.

Shared improvement also requires retaining useful behavior as new capabilities are acquired. Replacing the common harness with one successful specialist can lose capabilities developed elsewhere, while retaining specialists behind a router preserves separate solutions without necessarily combining their mechanisms. Optimizing directly on pooled experience provides access to broader feedback, but still requires reconciling workload-specific changes: an edit that benefits one workload may impair behavior needed by another. The challenge is thus to connect the discovery, retention, and dissemination of capabilities, so that progress across different workloads contributes to a shared agent rather than remaining fragmented or being lost through interference.

We introduce \textbf{EvolveNet}, a paradigm for \emph{collaborative harness evolution} that turns experience distributed across users into a shared process of agent self-improvement. EvolveNet is designed to expand exploration through parallel specialist search, accumulate useful capabilities while limiting destructive interference, and broadcast integrated discoveries to participating clients. Its central goal is a progressively stronger shared agent, not merely a collection of independently improved specialists. Each round begins with clients evolving the same broadcast harness on their respective workloads in parallel. Concurrent execution reduces the latency of the local-search phase relative to executing the same searches sequentially, while specialization directs the searches toward different recurring failures. A server then composes the resulting program modifications into an updated shared harness. Once accepted and redistributed, this harness makes the integrated mechanisms available to every client and provides the common starting point for subsequent evolution. This \emph{evolve-locally-then-aggregate} cycle links parallel discovery with shared accumulation: local experience contributes to the common program, and the common program carries those contributions into further local searches.

Realizing this paradigm requires an aggregation operator for executable programs. The coordinate-wise averaging used for model updates in federated learning \citep{mcmahan2017fedavg,li2020fedprox} does not directly apply to source-code modifications, and syntactic compatibility does not ensure behavioral compatibility. A verification rule strengthened by one client may be removed by another; locally useful prompt edits can damage behavior elsewhere; and individually effective mechanisms can interfere through shared control flow or runtime state. Aggregation must therefore determine not only which modifications to retain, but where they should apply and how their combination affects existing capabilities.

To this end, we develop evidence-guided, scope-typed program aggregation to address this challenge. Program deltas express each client's modifications relative to the common broadcast harness, while execution traces and evaluation feedback guide their integration. The merger distinguishes globally shared mechanisms from home-scoped extensions activated under observable domain conditions. This allows different local mechanisms to be retained under their respective conditions rather than forcing every apparent conflict into a single-winner decision. Candidate merges are evaluated against the broadcast harness before adoption, assessing the behavior of the integrated program rather than relying on the benefits observed in individual clients alone.

Our theoretical analysis characterizes conditions under which program composition supports capability accumulation. We derive lower bounds on composition gains from contextual program interactions and provide sufficient conditions for outperforming the best whole-program candidate in a fixed comparison family. Under effective isolation of disjoint home scopes, we express the contribution of domain-specific extensions as a domain-weighted sum of their conditional gains, defined relative to the integrated shared mechanisms. A complementary finite-proposal construction illustrates how cross-domain regressions can obstruct pooled selection and identifies conditions under which specialization followed by composition achieves higher expected performance under a specified proposal model.

We evaluate EvolveNet across five settings spanning text-to-SQL, data-science coding, competitive programming, software engineering, and agentic workflows. The final shared harness achieves higher held-out performance than both the unevolved harness and the end-to-end select-best-client protocol in all five settings. Further aggregation comparisons and behavioral analyses provide evidence that program composition can retain and combine improvements discovered by different clients, supporting shared capability accumulation from distributed local search.

Our contributions are threefold:
\begin{itemize}
\item We introduce \textbf{EvolveNet}, a paradigm for collaborative harness evolution that enables multiple clients to jointly improve a shared agent through local experience. By coupling parallel specialization with iterative program aggregation, it supports the accumulation of complementary capabilities in a shared harness and their dissemination across clients for subsequent evolution.

\item We develop an evidence-guided, scope-typed aggregation operator to integrate independently evolved mechanisms into a shared executable harness. Our theoretical analysis provides sufficient conditions for retaining complementary gains despite program interactions and for composition to improve over available whole-program candidates.

\item We evaluate EvolveNet across five settings, demonstrating higher held-out performance than unevolved and select-best-client baselines. Aggregation comparisons, behavioral analyses, and cost measurements examine capability accumulation and the computational trade-offs of collaborative evolution.
\end{itemize}

\section{Preliminaries}
\label{sec:prelim}

\paragraph{Harness and evaluation.}
Let $M$ be a frozen LLM and $\mathcal H$ the space of admissible
harnesses. A harness $h\in\mathcal H$ is an executable program that
constructs requests to $M$, invokes tools, processes their outputs,
and returns an answer to a task input $x$.
An evaluation item $z$ contains the input $x$ and the information
required for task-specific scoring. We use $s(h;z)$ to denote the
expected score of $h$ on $z$, averaging over execution randomness.

\paragraph{Distributed setting.}
We consider $K$ clients, where client $k$ holds a local workload
$D_k^{\mathrm{tr}}$ associated with domain $d_k$.
Clients evolve a common harness on their respective workloads and
return their programs and execution evidence to a server for
aggregation. A separate validation set $D^{\mathrm{val}}$,
disjoint from the local training workloads and final test set,
is used to assess aggregated candidates.
Labels are available for local candidate selection and server
validation; final performance is evaluated after the shared
harness is fixed.

\paragraph{Target distribution.}
Let $\mathcal P_k$ denote client $k$'s local workload distribution.
The client mixture is
$\mathcal P_{\mathrm{client}}
=\sum_{k=1}^{K}\pi_k\mathcal P_k$,
where $\pi_k\geq 0$ and $\sum_{k=1}^{K}\pi_k=1$.
The shared harness is evaluated on a target workload distribution
$\mathcal P_{\mathrm{target}}$, which need not coincide with the
client mixture and may include related domains absent from local
training workloads.

\begin{figure*}[t]
\centering
\vspace{-0.3cm}
\includegraphics[width=0.9\textwidth]{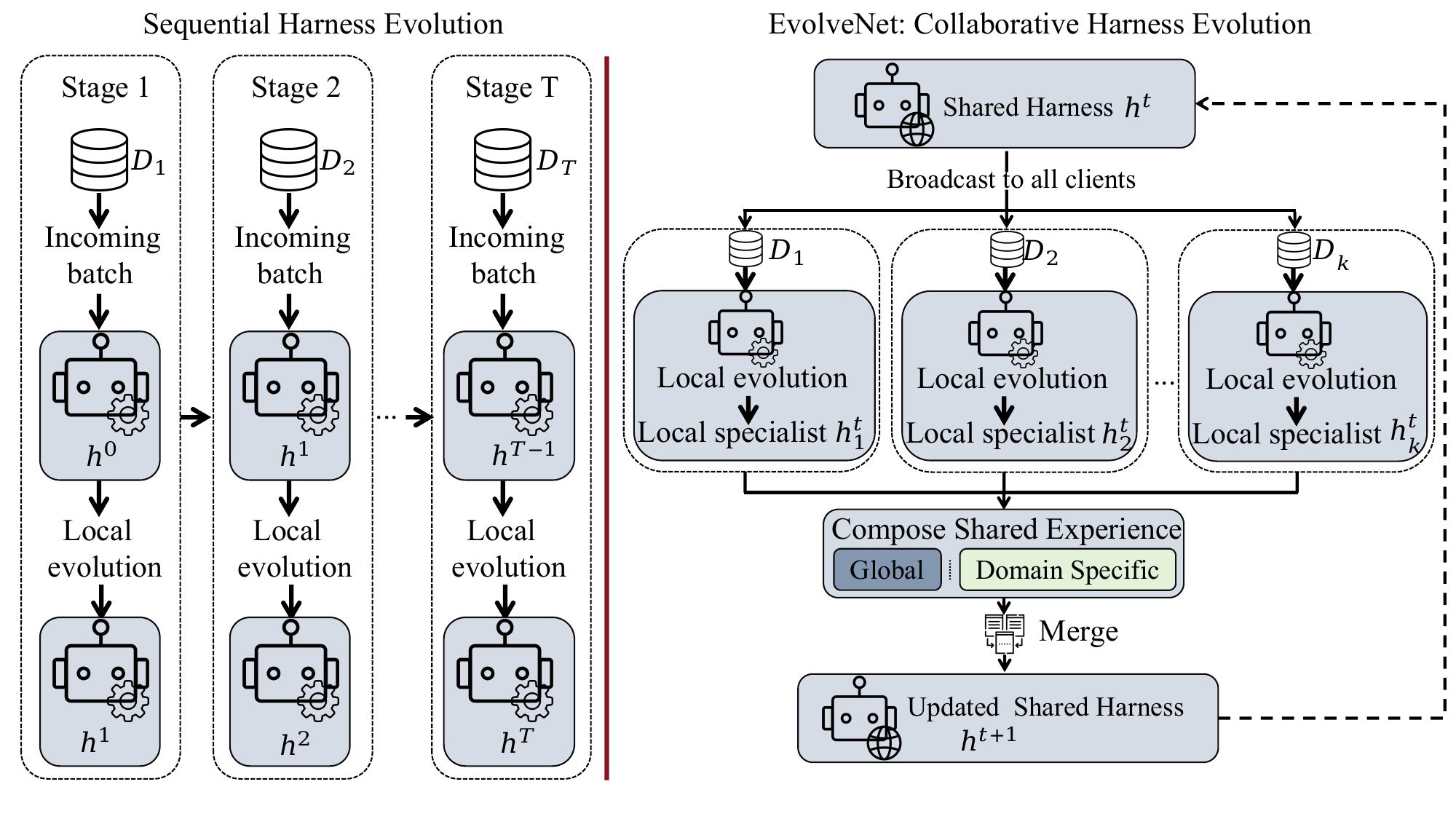}
\vspace{-0.3cm}
\caption{Sequential harness evolution (left) and EvolveNet (right).
EvolveNet combines parallel local specialization with program
aggregation and redistribution to evolve a shared harness.}
\vspace{-0.3cm}
\label{fig:method}
\end{figure*}

\section{Method: EvolveNet}
\label{sec:method}

EvolveNet aims to turn improvements discovered across client workloads
into cumulative progress in a single shared harness. Its target
objective is
\begin{equation}
h^\star \in \arg\max_{h\in\mathcal H}
\mathbb E_{z\sim\mathcal P_{\mathrm{target}}}[s(h;z)].
\label{eq:objective}
\end{equation}
The central question is how local evolution can contribute to this
shared objective when clients encounter different tasks and failure
modes. We address it at two levels: organizing local searches so that
their discoveries can be inherited across clients, and composing
those discoveries without treating every locally useful modification
as globally applicable.

\subsection{Collaborative harness evolution}
\label{sec:collaborative_evolution}

Execution-driven agent evolution improves programs through repeated
revision and evaluation. TTHE evolves multiple harness branches from
execution feedback and selects one to carry forward
\citep{nie2026tthe}; the Darwin G\"odel Machine maintains an archive
from which agents are selected for further self-modification
\citep{zhang2025dgm}. These mechanisms support exploration and
reuse of previous discoveries. EvolveNet addresses a complementary
problem: how searches driven by different client workloads can
jointly improve the program from which all clients continue evolving.

Applying candidate search directly to this setting creates a
distinction between local and shared progress. Selecting a program
on one client's workload does not establish its value for other
clients. Evaluating complete candidates on a pooled workload can
instead reject a locally useful change when its off-domain
regressions outweigh its local gains. In either case, the usefulness
of a mechanism is entangled with the performance of the complete
program that contains it. A mechanism may therefore be worth
retaining even when its originating specialist is not an appropriate
replacement for the shared harness. EvolveNet separates these decisions. At round $t$, client $k$
evolves the common broadcast harness $h^{(t)}$ on
$D_k^{\mathrm{tr}}$ to obtain a specialist $h_k^{(t)}$.
Its proposer follows trace-driven harness revision, with instructions to address recurring
structures and failures in the local workload. The broadcast
harness remains among its local candidates. Local selection
identifies promising behavior on that workload; server-side
aggregation determines how this behavior can be incorporated
into a program serving multiple workloads. A client thus
contributes potential mechanisms, rather than having to produce
a complete program that is already suitable for everyone.

This division also changes how searches interact. Clients explore
their workloads concurrently from the same shared state, directing
their searches toward different local problems. The local-search phase takes the maximum
rather than the sum of the client runtimes; composition and
validation remain additional costs.
After aggregation and validation, the committed harness
$h^{(t+1)}$ is broadcast to every client. Unlike continuing
independent specialist histories, each subsequent search can now
use mechanisms incorporated from other clients' discoveries. Algorithm~\ref{alg:fedhc} specifies this
\emph{specialize--aggregate--redistribute} cycle.

\subsection{Scope-typed program aggregation}
\label{sec:aggregation}

Parallel discovery alone does not produce a stronger shared program.
Selecting one specialist can omit useful mechanisms found by others,
while combining their source modifications may introduce
conflicts. The aggregation problem is therefore to preserve useful
differences where specialization matters, while integrating reusable
mechanisms into shared behavior.

\paragraph{Preserving specialization through scope.}
A locally beneficial modification need not be appropriate for every
input. Two clients may, for example, change the same verification
rule in opposite ways because their workloads require different
behavior. Selecting one implementation globally sacrifices the
other; applying both without conditions may make them conflict.
EvolveNet instead treats applicability as part of the adoption
decision, allowing a mechanism to be retained without requiring
it to act on every workload.

Let $\mathcal M=\{m_1,\ldots,m_n\}$ denote the candidate mechanisms
inferred from the clients' modifications. A mechanism is a
semantically related group of edits implementing a recognizable
behavior, potentially spanning multiple diff hunks.
The merger assigns
\begin{equation}
\tau(m)\in
\{\textsc{global}\}
\cup
\{\textsc{home}(d_k)\}_{k=1}^{K}
\cup
\{\bot\},
\label{eq:scope_type}
\end{equation}
where $\bot$ denotes rejection.
It is instructed to reject item-specific answer encodings,
incorporate mechanisms with cross-domain support or a
domain-independent rationale as shared behavior, and retain
domain-specific mechanisms under observable home-domain conditions.

Scope conditioning changes which conflicts require choosing a
winner. Modifications intended for disjoint domains can coexist
when their effects are confined to their respective conditions.
This requires the realized program to preserve dispatch, guard
the relevant effects, and prevent shared-state leakage
(Appendix~\ref{app:scope_isolation}). Interactions within the same
active region or through shared behavior still require reconciliation.
A general merger may also discover conditional code; EvolveNet
makes the distinction between shared and domain-specific behavior
an explicit part of its integration strategy. The result is not merely a router over complete specialists.
A shared recovery routine and a domain-conditioned check can
operate within the same control flow, allowing an execution to
use mechanisms originating from different clients. The purpose
of scoping is thus to preserve useful specialization within a
common program, rather than to keep all specialist programs
separate.

\paragraph{Grounding integration in client experience.}
Deciding where a mechanism should apply requires more than knowing
which client produced it. A client-specific program may contain
both generally useful procedures and domain-dependent rules.
EvolveNet therefore grounds scope assignment and integration in
the relationship between source modifications and observed behavior. Because all clients begin from $h^{(t)}$, client $k$'s contribution
has a common reference:
\begin{equation}
\Delta_k^{(t)}
=
\operatorname{diff}(h^{(t)},h_k^{(t)}).
\label{eq:program_delta}
\end{equation}
This distinguishes current modifications from inherited code.
The merger can reason about what each client changed without
treating the entire specialist program as a new contribution.

Execution evidence then informs the interpretation of these
changes. For binary outcomes, let $v_i(h)\in\{0,1\}$ denote
the recorded evaluator verdict on item $i$. Client $k$ reports
\begin{align}
F_k^{(t)}
&=
\{i\in D_k^{\mathrm{tr}}:
v_i(h^{(t)})=0,\,
v_i(h_k^{(t)})=1\},\\
B_k^{(t)}
&=
\{i\in D_k^{\mathrm{tr}}:
v_i(h^{(t)})=1,\,
v_i(h_k^{(t)})=0\}.
\end{align}
Together with the full execution traces $\mathcal T_k^{(t)}$,
these form
$e_k^{(t)}=(\mathcal T_k^{(t)},F_k^{(t)},B_k^{(t)})$.
The traces expose the execution contexts in which a modification
operates, while the verdict changes record the client variant's
observed gains and regressions. They support hypotheses about
useful mechanisms and their applicability, rather than establishing
an isolated causal effect for each edit.

The resulting aggregation operator is
\begin{equation}
\widetilde h^{(t+1)}
=
\mathcal A\!\left(
h^{(t)},
\{(\Delta_k^{(t)},e_k^{(t)})\}_{k=1}^{K}
\right).
\label{eq:aggregation_operator}
\end{equation}
The LLM merger integrates the inferred mechanisms by editing a
copy of $h^{(t)}$. Since this base carries previously adopted
behavior, the merger is instructed to prefer localized edits,
preserve existing prompt rules, and constrain code growth.
These instructions are intended to limit changes unrelated to
the new contributions, especially alterations to shared prompts
that can affect inputs outside the originating domain.

\paragraph{Validating shared updates.}
Because locally useful edits may interfere after composition, the
server evaluates $\widetilde h^{(t+1)}$ against $h^{(t)}$ on
$D^{\mathrm{val}}$ before redistribution. For binary outcomes,
a candidate is accepted if its measured validation accuracy is
no lower than that of $h^{(t)}$, including ties. Otherwise, the
merger revises it once using regression feedback, and the server
re-evaluates it. If the revision also fails, the server retains
$h^{(t)}$. The committed harness is redistributed as $h^{(t+1)}$
(Appendix~\ref{sec:gate}).

\newtheorem{theorem}{Theorem}
\newtheorem{lemma}[theorem]{Lemma}
\newtheorem{corollary}[theorem]{Corollary}

\section{Theoretical Analysis}
\label{sec:theory}

We analyze when separately discovered mechanisms can improve a shared
harness despite program interactions. Conditioning on a fixed
composition family, we derive composition-gain and marginal-retention
bounds, then specialize them to EvolveNet's shared mechanisms and
home-scoped extensions.

\paragraph{Executable composition family.}
Fix a broadcast program and mechanism blocks indexed by
$\mathcal M=\{1,\ldots,n\}$, with fixed implementations,
activation scopes, and insertion order.
For each $A\subseteq\mathcal M$, let $h_A$ denote the resulting
executable program, with $h_\varnothing$ the broadcast base.
Dependent edits are bundled so that every subset is well defined.
Let $q_A(z,\omega)\in[0,1]$ denote the realized task score of $h_A$.
Under a fixed evaluation law for $(Z,\Omega)$, define
\begin{equation}
J(A)=\mathbb E[q_A(Z,\Omega)].
\label{eq:theory_score}
\end{equation}
Here $X$ is the task-input component of $Z$, and $\Omega$ couples
executions from a common initial environment while preserving each
program's marginal execution law. Failures and budget exhaustion
receive defined scores. All gains use this same distribution;
scores may be binary or normalized continuous.

\paragraph{Contextual interaction.}
For distinct $i,j\in\mathcal M\setminus A$, define
\begin{align}
g_i(A)&=J(A\cup\{i\})-J(A),\\
I_{ij}(A)&=g_i(A\cup\{j\})-g_i(A).
\label{eq:theory_interaction}
\end{align}
Positive interaction indicates complementarity; negative interaction
indicates reduced marginal benefit from redundancy or harmful
interference. Assume $\beta_{ij}=\beta_{ji}\geq0$ such that
$I_{ij}(A)\geq-\beta_{ij}$ for every
$A\subseteq\mathcal M\setminus\{i,j\}$.
The bound is uniform over backgrounds and does not assume
pairwise-additive effects.

A measurable input region $Q_i$ is an effective activation region if
$q_{A\cup\{i\}}(Z,\Omega)=q_A(Z,\Omega)$ almost surely on
$\{X\notin Q_i\}$, for every
$A\subseteq\mathcal M\setminus\{i\}$.
The isolation conditions in
Propositions~\ref{prop:isolation}--\ref{prop:conflict}
suffice when they hold uniformly over the composition family.

\begin{lemma}[Scope overlap controls contextual interaction]
\label{lem:theory_support}
Under the effective-activation property, for distinct $i,j$ and every
$A\subseteq\mathcal M\setminus\{i,j\}$,
\begin{equation}
|I_{ij}(A)|\leq 2\Pr(X\in Q_i\cap Q_j).
\label{eq:theory_overlap}
\end{equation}
Thus disjoint effective scopes have zero interaction, and
$\beta_{ij}=2\Pr(X\in Q_i\cap Q_j)$ is a valid uniform bound.
\end{lemma}

The pointwise mixed difference vanishes outside $Q_i\cap Q_j$
and has magnitude at most two. The overlap bound can be loose
for shared mechanisms.

\begin{theorem}[Composition gain and marginal retention]
\label{thm:theory_composition}
For disjoint $C,S\subseteq\mathcal M$, the contextual bounds imply
\begin{equation}
\begin{aligned}
J(C\cup S)-J(C)&\geq L(S\mid C),\\
L(S\mid C)&:=
\sum_{i\in S}g_i(C)
-\sum_{\substack{i<j\\i,j\in S}}\beta_{ij}.
\end{aligned}
\label{eq:theory_gain}
\end{equation}
For disjoint old and new sets $O,N\subseteq\mathcal M$, they also imply
\begin{align}
&[J(O\cup N)-J(N)]-[J(O)-J(\varnothing)]\nonumber\\
&\hspace{10mm}\geq
-\sum_{i\in O}\sum_{j\in N}\beta_{ij}.
\label{eq:theory_retention}
\end{align}
The second bound controls the loss of the old blocks' marginal
contribution, not the number of previously solved items.
\end{theorem}

Composition improves on $C$ when the added mechanisms' contributions
on that background exceed the interaction penalty.
The proof telescopes contextual marginal gains without assuming
monotonicity, submodularity, or independent mechanism effects
(Appendix~\ref{app:theory}).

\paragraph{Composition versus whole-program selection.}
Fix $U\subseteq\mathcal M$ and a nonempty collection
$\mathcal C\subseteq 2^U$ of available candidate subsets.
Let $J_\star=\max_{C\in\mathcal C}J(C)$.
Theorem~\ref{thm:theory_composition} gives
\begin{equation}
\begin{aligned}
J(U)-J_\star
\geq \max_{C\in\mathcal C}\bigl\{
&J(C)-J_\star\\
&+L(U\setminus C\mid C)\bigr\}.
\end{aligned}
\label{eq:theory_compare}
\end{equation}
A positive right-hand side certifies an advantage over the best
whole program in this fixed family, including candidates with
multiple mechanisms or conditional behavior. The comparison
does not extend to input-wise oracle routing.
Appendix~\ref{app:theory} treats imperfect source integration
and candidate mismatch.

\paragraph{Shared mechanisms and home-scoped extensions.}
Let $C$ index the integrated shared mechanisms and $S_d$ the
home-scoped blocks for domain $d$.

\begin{corollary}[Accumulation on a shared background]
\label{cor:global_home}
Let $C,S_1,\ldots,S_D\subseteq\mathcal M$ be mutually disjoint,
and let $R_1,\ldots,R_D$ be disjoint measurable input regions.
Each block indexed by $S_d$ is effectively inactive outside $R_d$
for every admissible background.
Write $S=\bigcup_{d=1}^{D}S_d$ and $\alpha_d=\Pr(X\in R_d)$.
For $\alpha_d>0$, define
\begin{equation}
\delta_d(C)
=
\mathbb E[q_{C\cup S_d}-q_C\mid X\in R_d],
\label{eq:home_conditional_gain}
\end{equation}
where $q_A$ abbreviates $q_A(Z,\Omega)$. Then
\begin{equation}
J(C\cup S)-J(C)
=
\sum_{d:\alpha_d>0}\alpha_d\delta_d(C).
\label{eq:global_home_accumulation}
\end{equation}
\end{corollary}

Under effective isolation, home-scoped contributions add across
disjoint domains, while shared--local interactions remain in
$\delta_d(C)$. These conditional gains are measured after integration
with $C$ and may have either sign.
Appendix~\ref{app:global_home} gives the proof, realization-error
bound, and treatment of uncovered domains.
Proposition~\ref{prop:theory_pool} in Appendix~\ref{app:theory}
gives a finite-proposal example of interference-induced stagnation
in pooled search.

\section{Experiments}
\label{sec:experiments}

\subsection{Experimental setup}
\label{sec:protocol}

We evaluate EvolveNet on BIRD
\citep{li2023bird}, DS-1000 \citep{lai2023ds1000},
SWE-bench Verified \citep{jimenez2024swebench},
LiveCodeBench \citep{jain2025livecodebench}, and ClawEval.
Client workloads are partitioned by database, library, repository,
difficulty band, and task family, respectively.
We report accuracy for the first four settings and mean
task score for ClawEval.

Unless otherwise specified, deepseek-v4-flash serves as the frozen
solver, local proposer, and server merger. We use $T=3$ rounds
and $K=5$ clients ($K=3$ on LiveCodeBench), with one local
generation and one proposer branch per client per round.
Training, validation, and test sets are disjoint, and test
performance is measured after the evaluated harness is fixed.
BIRD is used for development; the aggregation operator is then
held fixed on the other four settings.
Benchmark construction, data splits, prompts, and measurement
details are provided in
Appendices~\ref{app:audit}--\ref{app:measurement}.

\subsection{Main results}
\label{sec:paired}

Table~\ref{tab:main} compares EvolveNet with the unevolved harness
and the end-to-end select-best baseline, which redistributes the
client program with the highest server-validation score after
each round of local evolution.

\begin{table*}[t]
\centering
\caption{Held-out performance (\%) across five benchmark subsets.
ClawEval reports mean task score; the other settings report accuracy.}
\label{tab:main}
\vspace{5pt}
\small
\begin{tabular}{lccccc}
\toprule
 & BIRD & DS-1000 & LCB & SWE-V & ClawEval \\
Test items & 150 & 200 & 30 & 40 & 30 \\
\midrule
Unevolved harness
& \num{57.3} & \num{55.5} & \num{33.3} & \num{37.5} & \num{65.8} \\
Select-best client
& \num{69.3} & \num{59.5} & \num{43.3} & \num{42.5} & \num{70.6} \\
\textbf{EvolveNet}
& \textbf{\num{70.7}} & \textbf{\num{68.5}} & \textbf{\num{66.7}}
& \textbf{\num{57.5}} & \textbf{\num{74.1}} \\
\bottomrule
\end{tabular}
\end{table*}

EvolveNet improves over the unevolved harness across all five settings,
demonstrating that collaborative harness evolution can translate
local experience into shared improvement without updating model
weights. It also outperforms select-best in every setting, including
gains of \num{9.0} percentage points on DS-1000 and \num{15.0}
on SWE-bench Verified. This advantage is consistent with complementary
improvements being distributed across clients: the highest-scoring
program need not contain every useful mechanism. Select-best can
therefore leave valuable discoveries in other clients unused,
whereas EvolveNet can integrate them under appropriate scopes,
allowing the shared harness to address failures that the selected
specialist leaves unresolved. Paired comparisons are provided in
Appendices~\ref{app:statistics} and~\ref{app:sb_all}. Figure~\ref{fig:rounds} shows that improvements occur at different
stages of evolution. BIRD gains mainly in the first round, whereas
DS-1000 and LiveCodeBench improve further in the second, showing
that subsequent rounds can improve an already evolved harness
on the validation workloads.

\begin{figure}[!htbp]
\centering
\begin{tikzpicture}
\begin{axis}[width=\linewidth, height=42mm, xlabel={EvolveNet round $t$},
  ylabel={Validation score (\%)},
  xmin=-0.15, xmax=3.15, ymin=30, ymax=98, xtick={0,1,2,3}, ytick={40,60,80},
  tick label style={font=\scriptsize}, label style={font=\scriptsize},
  axis line style={cbGray!60}, ymajorgrids, grid style={cbGray!20},
  legend style={font=\scriptsize, draw=none, fill=none, inner sep=1pt, row sep=-1pt,
                at={(0.02,1.02)}, anchor=north west, legend columns=2,
                /tikz/every even column/.append style={column sep=4pt}}]
\addplot[cbBlue, thick, mark=*, mark size=1.5pt]   coordinates {(0,56) (1,68) (2,68) (3,69)};
\addplot[cbOrange, thick, mark=square*, mark size=1.5pt] coordinates {(0,57) (1,61) (2,63) (3,63)};
\addplot[cbGreen, thick, mark=triangle*, mark size=1.8pt]  coordinates {(0,40) (1,50) (2,70) (3,70)};
\addplot[cbGray, thick, mark=diamond*, mark size=1.6pt] coordinates {(0,60) (1,70) (2,70) (3,70)};
\addplot[cbPurple, thick, mark=pentagon*, mark size=1.6pt] coordinates {(0,64.5) (1,72.8) (2,76.3) (3,76.3)};
\legend{BIRD, DS-1000, LCB, SWE, ClawEval}
\end{axis}
\end{tikzpicture}
\vspace{-0.1cm}
\caption{Validation performance of the committed shared harness
across rounds ($t=0$: unevolved).}
\label{fig:rounds}
\end{figure}
\vspace{-0.1cm}

\subsection{Comparing aggregation strategies}
\label{sec:aggablation}

We compare aggregation strategies on DS-1000 through end-to-end
evolution and replays from identical final-round client programs
(Table~\ref{tab:agg}). Each merger replay uses one realization;
both panels report the same final EvolveNet harness.

\paragraph{End-to-end comparison.}
Holistic aggregation uses execution traces to combine client
modifications and permits domain-conditioned code, but does not
require GLOBAL/HOME classification
(Appendix~\ref{app:prompts}).
Applied throughout evolution, it reaches \num{61.0}\%,
compared with \num{68.5}\% for EvolveNet.
The result supports explicit scope assignment as a guide to
retaining and integrating complementary local changes, beyond
merely permitting domain-conditioned code.

\begin{table*}[t]
\centering
\caption{DS-1000 aggregation comparisons (held-out accuracy, \%).
Left: end-to-end evolution.
Right: aggregation from an identical final-round client snapshot.}
\label{tab:agg}
\vspace{5pt}
\small
\begin{tabular}{lc@{\hspace{3em}}lc}
\toprule
\multicolumn{2}{c}{End-to-end evolution}
& \multicolumn{2}{c}{Fixed-client-snapshot aggregation} \\
\cmidrule(r){1-2}\cmidrule(l){3-4}
Aggregation rule & Accuracy & Aggregation rule & Accuracy \\
\midrule
Select-best client & \num{59.5}
& Delegation & \num{64.0} \\
Holistic aggregation & \num{61.0}
& GLOBAL-only merge & \num{66.5} \\
\textbf{EvolveNet} & \textbf{\num{68.5}}
& \textbf{EvolveNet} & \textbf{\num{68.5}} \\
\bottomrule
\end{tabular}
\end{table*}
\paragraph{Fixed-snapshot comparison.}
Using the same final-round client programs, we compare delegation,
which retains complete specialists and dispatches tasks by library,
with GLOBAL-only composition, which combines mechanisms without
domain conditioning. GLOBAL-only reaches \num{66.5}\%, exceeding
delegation's \num{64.0}\% even though domain conditions are forbidden.
This ordering supports a benefit from composition beyond specialist
routing: delegation restricts each task to the mechanisms within
one client program, whereas composition can integrate contributions
from different clients into the same execution procedure.

EvolveNet combines shared mechanisms with domain-scoped extensions
and reaches \num{68.5}\%. Its stronger performance suggests that
cross-client reuse and selective specialization can complement
one another: useful mechanisms can be shared without requiring
every retained modification to apply globally. The library-level breakdown illustrates this distinction
(Table~\ref{tab:bylib}). On Sklearn, delegation scores
\num{62.5}\%, GLOBAL-only composition \num{45.8}\%, and
EvolveNet \num{58.3}\%. This partial recovery is consistent with
scoped composition preserving useful domain-specific behavior
alongside shared mechanisms.

\subsection{Experience accumulation across agents}
\label{sec:retention}

We examine whether improvements discovered by different clients
are preserved in the shared harness. A client's \emph{gain} comprises
the held-out items it newly solves relative to that round's broadcast
harness. This reference excludes successes already present in the
shared base, focusing the analysis on discoveries from the current
local searches. On DS-1000, the five clients' gains cover \num{20}
distinct items, of which the merged harness retains \num{18}
(\num{90.0}\%). Thus most observed local gains survive their
integration into a common program.

The distribution of these gains reveals complementary coverage
across clients. Within the union, nine items are solved exclusively
by the Pandas client, six by Sklearn, and two by Matplotlib.
These disjoint groups leave only three other items, so any individual
client covers at most \num{12} items in this union, compared with
the merge's \num{18}. The shared harness therefore preserves a
broader set of newly acquired successes than any single specialist
provides. It also solves two test items missed by every individual
client. Together, these results show that the merged program retains
complementary client successes while extending beyond their
recorded coverage.

The shared harness improves over the unevolved program on every
DS-1000 library (Table~\ref{tab:bylib}), including Pytorch, which
appears in server validation and test but not in local training.
EvolveNet solves \num{10} of its \num{15} test items, compared
with \num{8} for both the unevolved harness and select-best.
Delegation and GLOBAL-only also solve \num{10}.
This observation complements the retention analysis: improvement
is not confined to domains with dedicated local specialists.

\paragraph{What is composed?}
The merge reports illustrate how local discoveries are incorporated
into shared behavior. On BIRD, the merger generalizes a
client-specific value-grounding probe and integrates another
client's SQL-dialect repair into a simpler retry structure.
The contribution is therefore not confined to copying the original
specialist procedure: its implementation can be revised to support
broader use within the shared program.

On DS-1000, cross-library execution-retry logic becomes shared
behavior, while Pandas schema pre-analysis remains conditional.
These cases illustrate two complementary forms of accumulation:
broadening the reuse of transferable mechanisms and retaining
specialized procedures under their applicable conditions.
This shared-background/local-extension structure corresponds to
the setting analyzed in Corollary~\ref{cor:global_home}.
Appendix~\ref{sec:qualitative} provides the detailed cases and
rejected rules.

\subsection{Comparison with centralized evolution}
\label{sec:central_main}

On BIRD's pooled \num{100}-item workload, centralized evolution
reaches \num{68.7}\% with one proposer per round and \num{67.3}\%
with five branches, compared with EvolveNet's \num{70.7}\%.
All run for three rounds; the five-branch baseline matches
EvolveNet's fifteen local proposer sessions, with aggregation
remaining an additional cost for EvolveNet.
The centralized population does not improve beyond its first-round
best, and recorded programs include a mechanism that helps one
database family while misleading others. These observations suggest
a difficulty of pooled selection: useful local changes can be
discarded together with their off-domain regressions.
Specialization followed by scoped composition offers a way to
retain such changes without applying them everywhere
(Appendix~\ref{sec:central}).

\subsection{Model-stack robustness and search cost}
\label{sec:modelablation_main}

\paragraph{A second model stack.}
Replacing the solver, proposer, and merger with MiMo-V2.5 under
the same BIRD protocol yields \num{72.7}\%, versus \num{62.7}\%
for the unevolved harness and \num{72.0}\% for select-best.
Thus substantial improvement through evolution also occurs with
another model stack, while composition provides a smaller
additional gain over selection in this run
(Appendix~\ref{sec:modelablation}).

\paragraph{Client count and parallel search.}
We repartition the same BIRD workload across $K{=}1,2,5,7$,
keeping rounds and per-client search budgets fixed.
Figure~\ref{fig:kscaling} shows that accuracy peaks among these
configurations at \num{70.7}\% for $K{=}5$.
Increasing to $K{=}7$ raises round-level speedup from
\num{1.81}$\times$ to \num{2.22}$\times$ but yields
\num{70.0}\% accuracy. Scheduling gains therefore need not
translate into better final programs; these configurations also
differ in workload partitioning and total proposer sessions.

At $K{=}5$, local searches achieve \num{3.96}$\times$ speedup
over their serial execution, reduced to \num{1.81}$\times$
after including aggregation and validation. Integration therefore
limits how much local parallelism accelerates the complete round.
The full run takes \num{6110}\,s, versus \num{1693}\,s and
\num{3655}\,s for centralized single-branch and population search,
respectively: the observed accuracy advantage comes with a
longer end-to-end runtime (Appendix~\ref{sec:cost}).

\begin{figure}[t]
\centering
\begin{tikzpicture}
\begin{axis}[width=\linewidth, height=48mm,
  xlabel={Round speedup vs.\ serial ($\times$)},
  ylabel={BIRD held-out (\%)},
  xmin=0.85, xmax=2.45, ymin=67.8, ymax=71.4, xtick={1,1.5,2}, ytick={68,69,70,71},
  tick label style={font=\scriptsize}, label style={font=\scriptsize},
  axis line style={cbGray!60}, ymajorgrids, grid style={cbGray!20}]
\addplot[cbBlue, thick, mark=*, mark size=2pt] coordinates {(1.00,68.7) (1.33,69.3) (1.81,70.7) (2.22,70.0)};
\node[font=\scriptsize, anchor=north west] at (axis cs:1.02,68.7) {$K{=}1$};
\node[font=\scriptsize, anchor=north west] at (axis cs:1.35,69.3) {$K{=}2$};
\node[font=\scriptsize, anchor=south east] at (axis cs:1.79,70.8) {$K{=}5$};
\node[font=\scriptsize, anchor=north east] at (axis cs:2.20,69.9) {$K{=}7$};
\end{axis}
\end{tikzpicture}
\caption{BIRD held-out accuracy and round-level speedup at different
client counts. Speedup is relative to serial execution of the same
local searches at each $K$, with aggregation and validation included
in both schedules.}
\label{fig:kscaling}
\end{figure}

\section{Conclusion}

We introduced EvolveNet, a framework for collaborative harness
evolution that connects parallel local specialization with
evidence-guided, scope-typed program aggregation and redistribution.
Across five settings, EvolveNet improves held-out performance over
both the unevolved harness and the end-to-end select-best baseline.
Aggregation comparisons and behavioral analyses support the retention
and combination of complementary client improvements within a single
shared program. Our theoretical analysis provides sufficient conditions
for retaining and combining such gains despite program interactions.
Together, these findings support executable harnesses as a medium for
collaborative self-improvement, allowing discoveries from different
clients to become a shared basis for subsequent evolution.

\clearpage
\bibliographystyle{plainnat}
\bibliography{refs}

\clearpage
\onecolumn
\appendix
\section{Benchmark Construction and Gold Auditing}
\label{app:audit}

\paragraph{Per-item gold verification.} Every training, validation, and test item passed a benchmark-specific
gold check in our execution environment before use. For DS-1000, each problem's reference solution was
executed against its own hidden test; \num{39} of \num{400} initially sampled problems failed (predominantly
Matplotlib, Tensorflow and Pytorch environment sensitivities) and were replaced by same-library problems
that pass, topping up from the four largest libraries once the Tensorflow/Pytorch clean pools were
exhausted. For SWE-bench Verified, each instance's gold patch was run through the official evaluation
harness in our Docker environment; all \num{80} selected instances resolve. For LiveCodeBench, all
\num{110} selected problems carry non-empty test suites.

\paragraph{BIRD gold audit.} BIRD gold queries are known to contain defects. An LLM auditor reviewed every
held-out item under a deliberately conservative policy whose default is \emph{accept}; an item is rejected
only if (i) the gold SQL fails to execute, (ii) the hint contradicts the question on a concrete value,
number, or named entity, or (iii) the gold silently drops a stated condition or adds an unstated one.
Difficulty is never grounds for rejection: case mismatches between hint and stored data, discretionary
\texttt{DISTINCT}, and projection choices are the solver's job and are explicitly accepted. The audit
is applied during construction of the held-out set, which contains \num{150} items: \num{40} hard and
\num{110} representative. All reported BIRD held-out results use this test set.
We validated the criteria by falsification: an earlier, stricter version rejected \num{60}\% of a
human-audited slice, and items it flagged were solved at the same rate as items it passed --- i.e., it had
no discriminative power --- which forced the weaker, execution-grounded criteria above.

\paragraph{Audit inputs and development history.}
The auditor receives the database schema, question, hint, gold SQL,
and gold-query execution result. Per-item audit decisions use these
inputs without access to harnesses, generated SQL, or system scores.
The criteria were calibrated using solve rates from earlier
exploratory runs. The final criteria and item list were frozen on
2026-07-22, before the reported end-to-end experiments; the earliest
BIRD run completed on 2026-07-23. The item list was not subsequently
revised. The audit prompt, criteria, and excluded-item list are
released with the code. No independent human audit was conducted.

\section{Shards and Splits}
\label{app:shards}

\paragraph{Evaluation workloads.}
ClawEval contains agentic workflow tasks over an OpenClaw agent with
mock enterprise services. LiveCodeBench serves as a smaller robustness
check since its single-platform pool admits no five-way sharding.
The client-mixture weights $\pi_k$ are uniform. Held-out sets contain
\num{150} items on BIRD, \num{200} on DS-1000, \num{40} on SWE-bench
Verified, and \num{30} each on ClawEval and LiveCodeBench.

All splits are drawn with a fixed seed (0) and are disjoint by construction. BIRD: five clients hold
\num{24}/\num{18}/\num{18}/\num{18}/\num{22} training items covering card\_games, california\_schools,
formula\_1, toxicology, and a mixed shard (debit\_card\_specializing, superhero, student\_club); validation
(\num{100}) and test (\num{150}) are stratified over eleven databases, four of which have no client
specialist. DS-1000: five clients hold \num{20} items each from Pandas, Numpy, Matplotlib, Sklearn, and
Scipy; validation and test are stratified over the remaining pool, so Pytorch appears only as a
specialist-free domain; Tensorflow items did not survive the gold check and were replaced from the
larger libraries, leaving six libraries in the final splits. SWE-bench Verified: five clients hold \num{30} items across django, sphinx,
sympy, astropy, and a pytest+pylint shard, bounded by locally available evaluation images; the test set
adds \num{10} instances from repositories no client trained on. LiveCodeBench: three clients hold \num{20}
items per difficulty band. ClawEval: five clients hold \num{8}/\num{8}/\num{4}/\num{4}/\num{6} training
items drawn from the workflow, operations, communication, productivity, and miscellaneous task families,
with validation (\num{10}) and test (\num{30}) allocated from the same families. The $K$-ablation regroups the BIRD shards ($K{=}2$) or splits the mixed shard
by database ($K{=}7$) with all other settings unchanged.

\section{Prompts}
\label{app:prompts}

\paragraph{Local search configuration.}
Each client's proposer is given a search role---conservative repair,
independent exploration, or adversarial audit---assigned by client
index so that concurrent branches do not collapse onto the same edit.
We use $K{=}5$ clients except on LiveCodeBench ($K{=}3$), and $T{=}3$
rounds, with one local generation and one proposer branch per client
per round ($E{=}G{=}1$ in Algorithm~\ref{alg:fedhc}).

\paragraph{Specialist instruction (abridged).} Appended to every client proposer prompt:
\begin{quote}\small
``SPECIALIST MODE --- you serve ONE client whose data is the slice in these traces (often a single
database). Do NOT aim for a generic harness that works everywhere; aim to become the EXPERT of THIS slice.
Study the traces for the recurring, slice-specific failures [\dots] Encode that specific knowledge ---
table/column disambiguation, value formats, join paths particular to THIS schema --- as durable mechanisms.
(Still no hardcoding one question's literal answer; schema- and slice-level knowledge is exactly what you
SHOULD capture.) The server will later merge your expertise with other clients' --- the more your harness
knows that theirs cannot, the more the collaboration gains.''
\end{quote}
The DS-1000, SWE-bench, and LiveCodeBench variants substitute library, repository, and difficulty-band
wording respectively.

\paragraph{Scope-typed adoption rule (abridged).} The aggregator receives, in addition to the shared
merge scaffold (deltas against the broadcast base, proposal cards, graded verdicts, line budget):
\begin{quote}\small
``SPLIT EVERY CANDIDATE MECHANISM INTO GLOBAL vs HOME-SCOPED before deciding. GLOBAL: mechanisms whose
evidence spans databases or whose failure mode is universal [\dots] adopt ONE best implementation globally.
HOME-SCOPED: a rule whose evidence comes from ONE client's home database [\dots] do NOT adopt it globally,
and do NOT reject it for regressing on OTHER databases --- adopt it CONDITIONALLY: the harness receives the
domain at solve time, so wrap the rule so it only applies on that client's home domain. A home-scoped rule
can only change behaviour at home, so off-home regressions are impossible by construction; judge it ONLY on
its home evidence. Still REJECT single-question hacks. CONFLICTS DISSOLVE UNDER SCOPING: when two clients
changed the same behaviour in opposite directions, keep each side scoped to its own home domain instead of
picking a winner.''
\end{quote}
The holistic, quorum, and conflict-resolution variants replace only this adoption block; scaffold,
evidence, and budget are identical across variants. The quoted instruction specifies the intended scope discipline; the sufficient conditions for its isolation claim are stated in Propositions~\ref{prop:isolation} and \ref{prop:conflict}, and are not established merely by including this instruction in a prompt.

\paragraph{Holistic aggregation.} The holistic rule uses execution traces to adopt, reject, or combine mechanisms, selecting a single preferred implementation among alternative solutions to the same problem. Unlike scope-typed aggregation, it does not require GLOBAL/HOME classification or explicit scope reporting, but it permits domain-conditioned branches. It therefore differs from GLOBAL-only aggregation, which forbids domain conditioning. The DS-1000 holistic result in Table~\ref{tab:agg} uses this rule throughout the collaborative loop, not only in a final-snapshot replay.

\section{Measurement Details}
\label{app:measurement}

\paragraph{Model and evaluation protocol.}
The frozen solver is deepseek-v4-flash, and the same model drives the
client proposers and server aggregator; Sec.~\ref{sec:modelablation_main}
repeats the pipeline with a different stack. Only the harness changes:
a static auditor rejects candidates it identifies as changing the
model, endpoint, or credentials, or reading a gold answer at inference.
Labels enter the loop at two points: clients select candidates on their
own shard, and the server gates merges on its validation slice. Test
sets are evaluated after the shared harness is frozen. Every item
passed a benchmark-specific gold or test-availability check in our
environment before use (Appendix~\ref{app:audit}).

\paragraph{Development protocol.}
BIRD served as the development benchmark, on which the aggregation
operator, the gate, and the loop design were selected. The operator
was then frozen and the remaining four settings were each run
end-to-end once. DS-1000 was designated in advance as the primary
confirmation benchmark, being the setting whose clients hold the
most disjoint workloads.

Two engineering measures improve repeatability. All frozen-solver replies are cached on the
full request (prompt, system message, temperature, sample count, and repeat index), so two harnesses that
ask the same question receive the same answer. Every child process runs with a pinned hash seed: without
it, a harness that collects column or table names in a set iterates them in a different order per process,
that order reaches the prompt, the cache key changes, and sampling variability reappears. Measured directly, the same frozen harness scored \num{16}/\num{17}/\num{17} on three identical
evaluations before the seed was pinned; afterwards, two full repeats produced byte-identical SQL on
\num{50}/\num{50} items.

End-to-end comparisons evaluate the complete evolution protocols.
Fixed-snapshot replays apply alternative aggregation rules to the
same client programs, controlling which discoveries are available
for composition. Each replay uses one realized merge. We summarize
paired outcomes on the same held-out items in
Appendix~\ref{app:statistics}. ClawEval uses continuous per-task
scores from an independent judge model.

\paragraph{ClawEval scoring and commitment.}
\label{app:claweval_gate}
ClawEval task scores are assigned by Gemini 2.5 Flash
(\texttt{gemini-2.5-flash}). On the fixed ten-task validation set,
let $f$ count tasks whose score increases by more than $0.05$
relative to the current shared harness, and let $b$ count tasks
whose score decreases by more than $0.05$. All other changes
are treated as unchanged. A candidate is accepted if
$f\geq b$ and $f+b>0$; otherwise, the current harness is retained.
Evaluation reports mean task scores, while commitment uses these
thresholded task-level counts.

\subsection{Paired outcome comparisons}
\label{app:statistics}

All comparisons use paired outcomes on the same held-out items.
For binary benchmarks, a win (loss) is an item solved only by
EvolveNet (only by the comparator). For ClawEval, the counts denote
improved and worsened continuous task scores. These summaries
compare the evaluated programs on the fixed held-out sets.

\paragraph{Comparison with the unevolved harness.}
The following counts correspond to Table~\ref{tab:main}.
\begin{center}\small
\begin{tabular}{lccccc}
\toprule
 & BIRD & DS-1000 & LCB & SWE-V & ClawEval \\
\midrule
Win / loss & 26 / 6 & 30 / 4 & 10 / 0 & 9 / 1 & 18 / 7 \\
\bottomrule
\end{tabular}
\end{center}
Appendix~\ref{app:sb_all} reports the paired counts for the
end-to-end select-best comparison.

\paragraph{Aggregation, centralized search, and the second model stack.}
The DS-1000 delegation comparison uses the fixed final-round client
snapshot of Table~\ref{tab:agg}. The BIRD population comparison
contrasts the complete protocols in Section~\ref{sec:central_main}.
MiMo-V2.5 replaces the solver, proposer, and merger as described in
Section~\ref{sec:modelablation_main}.
\begin{center}\small
\begin{tabular}{llc}
\toprule
Setting & Comparator & Win / loss \\
\midrule
DS-1000 & Delegation & 15 / 6 \\
BIRD & Centralized five-branch population & 14 / 9 \\
BIRD, MiMo-V2.5 & Unevolved harness & 23 / 8 \\
BIRD, MiMo-V2.5 & Select-best client & 10 / 9 \\
\bottomrule
\end{tabular}
\end{center}

\section{Keeping the Strongest Client, Across All Five Settings}
\label{app:sb_all}

Sec.~\ref{sec:aggablation} compares delegation, GLOBAL-only composition, and EvolveNet on one identical DS-1000 client snapshot. Separately, the end-to-end \emph{select-best} protocol promotes the client scoring highest on the server's validation slice without aggregation. It was run across the same five settings and held-out sets as Table~\ref{tab:main}:

\begin{center}\small
\begin{tabular}{lccccc}
\toprule
 & BIRD & DS-1000 & LCB & SWE-V & ClawEval \\
\midrule
Select-best client & \num{69.3} & \num{59.5} & \num{43.3} & \num{42.5} & \num{70.6} \\
\textbf{EvolveNet} & \textbf{\num{70.7}} & \textbf{\num{68.5}} & \textbf{\num{66.7}} & \textbf{\num{57.5}} & \textbf{\num{74.1}} \\
\midrule
paired win / lose & 9 / 7 & 33 / 15 & 7 / 0 & 8 / 2 & 17 / 8 \\
\bottomrule
\end{tabular}
\end{center}

EvolveNet achieves higher held-out scores than select-best in all
five settings. These end-to-end comparisons complement the
fixed-client-snapshot analysis in Section~\ref{sec:aggablation}.

\section{Gate-Rule Replay}
\label{app:gate}

Replaying every acceptance decision under a strict-improvement rule ($|\textit{fixed}|>|\textit{broke}|$,
ties rejected) leaves the committed artifact unchanged on BIRD and changes it on DS-1000 and
LiveCodeBench. For the runs whose strict-replay artifact was itself evaluated on held-out data:

\begin{center}\small
\begin{tabular}{lccc}
\toprule
Run & tie-accepting (protocol, \%) & strict replay (\%) & difference (pts) \\
\midrule
BIRD $K{=}5$ & \num{70.7} & \num{70.7} & --- \\
BIRD $K{=}2$ & \num{69.3} & \num{69.3} & --- \\
DS-1000 & \num{68.5} & \num{61.5} & $-7.0$ \\
LiveCodeBench & \num{66.7} & \num{50.0} & $-16.7$ \\
\bottomrule
\end{tabular}
\end{center}

In these replays, the tie-accepting artifacts solve \num{14} and \num{5} more held-out items on DS-1000 and LiveCodeBench, respectively. A validation tie can retain a program change with different behavior outside the validation slice. The replay is evidence about the recorded candidates and commitment decisions, not a replacement for independently rerunning all later local searches under the counterfactual gate.

\section{Related Work}

\paragraph{Prompt, context, and workflow optimization.}
LLM systems can improve by revising the structures around a frozen
model. Prompt optimization searches over instructions through
reflection, evolutionary search, and LLM-generated proposals
\citep{zhou2023ape,fernando2024promptbreeder,yang2024opro,guo2024evoprompt}.
GEPA uses trajectory-level reflection and combines complementary
prompt improvements from a Pareto frontier \citep{agrawal2026gepa},
while ACE maintains evolving contextual playbooks through generation,
reflection, and curation \citep{zhang2026ace}.
DSPy optimizes multi-stage pipelines \citep{khattab2024dspy}; ADAS
and AFlow search agent implementations and workflows
\citep{hu2025adas,zhang2025aflow}; and AgentSquare evolves and
recombines modules with standardized interfaces
\citep{shang2024agentsquare}.
Other approaches retain experience in different forms: Self-Refine
iteratively revises task outputs \citep{madaan2023selfrefine},
Reflexion records verbal feedback in episodic memory
\citep{shinn2023reflexion}, and Voyager builds a reusable library
of executable skills \citep{wang2024voyager}.
These approaches motivate distinguishing the artifact being improved
from the process used to improve it: outputs, contextual knowledge,
skill libraries, and complete control programs support different
forms of reuse and composition.

\paragraph{Harness and program self-evolution.}
Executable programs are themselves a search space for improvement.
STOP recursively improves code-generating programs
\citep{zelikman2023stop}, and FunSearch combines LLM proposals with
program evaluation \citep{romeraparedes2024funsearch}.
SICA, the Darwin G\"odel Machine, and MOSS explore agent
self-modification through source-code changes
\citep{robeyns2025sica,zhang2025dgm,cai2026moss}.
EvoTest, Meta-Harness, Self-Harness, and TTHE use execution feedback
to improve persistent agent configurations and harnesses
\citep{he2025evotest,lee2026metaharness,zhang2026selfharness,nie2026tthe}.
Recent systems extend this process in several directions.
DarwinX maintains alternative harness lineages and recombines
complementary variants under a preservation criterion
\citep{zhang2026darwinx}; SHE targets safety-related harness
artifacts using trajectory diagnoses \citep{qu2026she}; PILOT
couples live execution steering with updates to persistent skills
and memory \citep{xiao2026pilot}; and VERSE evolves both an
executor harness and its optimizer's procedures using
execution-based verification \citep{wang2026verse}.
Related analyses distinguish transferable harness mechanisms from
language-specific engineering knowledge \citep{yang2026onerecipe}
and study approximation, generalization, and optimization limits
of harness-based personalization \citep{gan2026harnesslearning}.
EvolveNet focuses on repeated integration and redistribution of
changes discovered on different client workloads from a common
broadcast base. Its scope assignments distinguish shared mechanisms
from domain-conditioned extensions, and its analysis studies their
interactions after executable composition.

\paragraph{Federated optimization and personalization.}
EvolveNet shares the local-update, aggregation, and redistribution
schedule of federated learning. FedAvg combines local parameter
updates \citep{mcmahan2017fedavg}; FedProx addresses heterogeneity
through proximal local objectives \citep{li2020fedprox}; SCAFFOLD
uses control variates to correct client drift
\citep{karimireddy2020scaffold}; and adaptive federated optimization
introduces adaptive server updates \citep{reddi2021fedopt}.
Personalized methods explicitly balance sharing and specialization.
FedPer and FedRep combine shared network components or
representations with client-specific layers or heads
\citep{arivazhagan2019fedper,collins2021fedrep}.
pFedMe and Ditto couple personalized models with global learning
through regularized objectives \citep{dinh2020pfedme,li2021ditto},
while IFCA learns cluster-specific models for groups of clients
\citep{ghosh2020ifca}.
EvolveNet addresses a related sharing--specialization tension in
program space: it integrates reusable and conditionally activated
mechanisms into one shared harness rather than aggregating model
weights or maintaining personalized parameter vectors.

\paragraph{Collaborative improvement through shared artifacts.}
Agent collaboration can exchange learned artifacts beyond full
model parameters. Fed-SE aggregates low-rank updates learned from
local trajectories \citep{chen2025fed}; FederatedSkill exchanges
semantic skill diffs to support personalized skill evolution
\citep{yang2026federatedskill}; and Federation over Text consolidates
local reasoning traces into a shared insight library
\citep{yao2026federation}.
EvolveNet instead composes client modifications into a shared
executable control program, where contributions can interact through
prompts, tool calls, control flow, and runtime state.
Its collaboration occurs through the program redistributed across
rounds. This differs from inference-time debate and multi-agent
conversation, which coordinate agents while solving a task
\citep{du2023debate,wu2023autogen}.

\paragraph{Model merging.}
Model merging also combines capabilities acquired from different
workloads. Task Arithmetic represents fine-tuning as parameter
displacements \citep{ilharco2023taskarithmetic}; TIES-Merging
reduces interference by resolving conflicting updates
\citep{yadav2023ties}; and DARE sparsifies and rescales parameter
deltas before merging \citep{yu2024dare}.
EvolveNet shares the objective of integrating improvements relative
to a common base, but source-code modifications do not directly admit
coordinate-wise arithmetic. It uses execution evidence and activation
scopes to guide integration, then validates the resulting program.

\section{Additional Method Details}
\subsection{From centralized evolution to collaborative harness evolution}
\label{sec:paradigm}

A centralized adaptation pipeline can first pool workloads and then evolve a harness:
\begin{equation}
\underbrace{\operatorname{Pool}(D_1^{\mathrm{tr}},\ldots,D_K^{\mathrm{tr}})\;\longrightarrow\;\operatorname{Evolve}}_{\text{centralized harness evolution}}.
\label{eq:centralized_paradigm}
\end{equation}
The optimizer accesses the pooled workloads and can use either a single proposer or a population, as in our centralized baselines.

EvolveNet instead adapts on each local workload before aggregating the resulting programs (Fig.~\ref{fig:method}):
\begin{equation}
\underbrace{\operatorname{Evolve}_{1:K}(h^{(t)},D_{1:K}^{\mathrm{tr}})\;\longrightarrow\;\operatorname{Aggregate}(\Delta_{1:K}^{(t)})}_{\text{collaborative harness evolution}}.
\label{eq:federated_paradigm}
\end{equation}
Client $k$ applies the local evolution operator $\mathcal E$ to the common broadcast harness:
\begin{equation}
h_k^{(t)}=\mathcal E(h^{(t)},D_k^{\mathrm{tr}}).
\label{eq:local_evolution}
\end{equation}
Its proposer is encouraged to specialize in recurring local structures and failures. Because the broadcast harness remains in the candidate pool, selection under the same recorded client score $S_k$ yields
\begin{equation}
S_k(h_k^{(t)})\ge S_k(h^{(t)}).
\label{eq:local_nondegradation}
\end{equation}
This is a property of empirical candidate selection under the measured score, not a population guarantee. Re-evaluating with fresh randomness can yield different scores.

The server integrates the returned adaptations into a candidate:
\begin{equation}
\widetilde h^{(t+1)}=\mathcal A(h^{(t)},h_1^{(t)},\ldots,h_K^{(t)}).
\label{eq:global_aggregation_highlevel}
\end{equation}
After validation, the committed program is redistributed. The resulting $\textsc{specialize}\rightarrow\textsc{aggregate}\rightarrow\textsc{redistribute}$ cycle makes an accepted local adaptation part of the starting state available to all clients in the next round. Clients do not exchange messages within a round; their local searches are coupled through the shared artifact across rounds.

Let $C_k^{(t)}$ denote local-search latency, $C_{\mathcal A}^{(t)}$ aggregation latency, and $C_{\mathrm{val}}^{(t)}$ validation latency. Executing the same local searches sequentially costs $\sum_k C_k^{(t)}+C_{\mathcal A}^{(t)}+C_{\mathrm{val}}^{(t)}$. With sufficient parallel resources and communication included in the corresponding phase costs, the parallel schedule is
\begin{equation}
C_{\mathrm{EvolveNet}}^{(t)}=\max_k C_k^{(t)}+C_{\mathcal A}^{(t)}+C_{\mathrm{val}}^{(t)}.
\label{eq:serial_depth}
\end{equation}
Equation~\ref{eq:serial_depth} compares parallel and serial schedules for the same local searches, including aggregation and validation costs.

\subsection{The program-aggregation challenge}
\label{sec:aggregation_challenge}

Program aggregation couples three questions. \emph{Attribution} asks which observed changes accompany a client variant and which mechanisms plausibly explain them; client-level evidence does not uniquely identify a causal source edit. \emph{Interference} concerns adaptations that conflict through control flow, prompts, runtime state, or domain assumptions. \emph{Regression} concerns capabilities lost after the adaptations are placed in the same executable program. A common base makes deltas comparable; scope assignments specify intended activation regions; and validation measures the behavior of the actual composition. These structures constrain the merger without making semantic compatibility decidable from a diff alone.

\subsection{Behavioral commitment and rollback}
\label{sec:gate}

EvolveNet separates construction of a candidate from commitment. For binary evaluation, let $D^{\mathrm{val}}$ be the fixed validation slice, disjoint from local training and final testing, and define
\begin{align}
F_{\mathrm{val}}&=\{i\in D^{\mathrm{val}}:v_i(h^{(t)})=0,\ v_i(\widetilde h^{(t+1)})=1\},\\
B_{\mathrm{val}}&=\{i\in D^{\mathrm{val}}:v_i(h^{(t)})=1,\ v_i(\widetilde h^{(t+1)})=0\}.
\end{align}
The candidate is accepted when
\begin{equation}
|F_{\mathrm{val}}|\ge|B_{\mathrm{val}}|.
\label{eq:acceptance_gate}
\end{equation}
The criterion is equivalent to non-decreasing accuracy on the
recorded validation outcomes, allowing gains to offset individual
regressions. Item-level transitions provide feedback for inspecting
and revising regressions.

When a candidate fails, the aggregator receives regression feedback
and may revise it once. A second failure causes rollback to
$h^{(t)}$, rather than promotion of a client program. Validation
outcomes are reused for revision and adoption across rounds;
final performance is evaluated on a separate test set.

With distributed binary validation, clients can return counts of
fixes and breaks for the server to sum; revision additionally
requires detailed regression feedback. Equation~\ref{eq:acceptance_gate}
and Algorithm~\ref{alg:fedhc} describe the binary-outcome gate.
ClawEval uses the thresholded task-count criterion in
Appendix~\ref{app:claweval_gate}.

\subsection{The complete EvolveNet procedure}
\label{sec:procedure}

\begin{algorithm}[t]
\caption{EvolveNet: one communication round (binary-outcome gate)}
\label{alg:fedhc}
\begin{algorithmic}[1]
\STATE \textbf{input:} shared harness $h^{(t)}$; client workloads $D_{1:K}^{\mathrm{tr}}$; validation
slice $D^{\mathrm{val}}$; local search budget $(E,G)$
\FOR{$k=1$ \TO $K$ \textbf{in parallel}}
    \STATE $h_k^{(t)} \leftarrow \textsc{LocalEvolve}\left(h^{(t)},D_k^{\mathrm{tr}},E,G\right)$
    \STATE $\Delta_k^{(t)} \leftarrow \operatorname{diff}\left(h^{(t)},h_k^{(t)}\right)$
    \STATE $(F_k^{(t)},B_k^{(t)}) \leftarrow \textsc{BehaviorChanges}\left(h_k^{(t)},h^{(t)};D_k^{\mathrm{tr}}\right)$
    \STATE $e_k^{(t)} \leftarrow (\mathcal T_k^{(t)},F_k^{(t)},B_k^{(t)})$, including the full local execution traces
\ENDFOR
\STATE $\widetilde{h}^{(t+1)} \leftarrow \textsc{ScopeAggregate}\left(h^{(t)},
        \left\{\left(\Delta_k^{(t)},e_k^{(t)}\right)\right\}_{k=1}^{K}\right)$
\STATE $(F_{\mathrm{val}},B_{\mathrm{val}}) \leftarrow \textsc{Compare}\left(\widetilde{h}^{(t+1)},h^{(t)};D^{\mathrm{val}}\right)$
\IF{$|B_{\mathrm{val}}|>|F_{\mathrm{val}}|$}
    \STATE $\widetilde{h}^{(t+1)} \leftarrow \textsc{ReviseAggregate}\left(h^{(t)},
            \left\{\left(\Delta_k^{(t)},e_k^{(t)}\right)\right\}_{k=1}^{K}, B_{\mathrm{val}}\right)$
    \STATE recompute $(F_{\mathrm{val}},B_{\mathrm{val}})$
\ENDIF
\STATE $h^{(t+1)} \leftarrow \widetilde{h}^{(t+1)}$ \textbf{if} $|F_{\mathrm{val}}|\ge|B_{\mathrm{val}}|$
        \textbf{else} $h^{(t)}$
\STATE \textbf{return} $h^{(t+1)}$
\end{algorithmic}
\end{algorithm}

Algorithm~\ref{alg:fedhc} states one round with the binary gate. The server broadcasts a common program; clients return evolved variants, full execution traces, and evaluation feedback; the merger composes a candidate; and validation determines commitment, revision, or rollback. Scope annotations guide composition, while Propositions~\ref{prop:isolation} and \ref{prop:conflict} specify the semantic premises needed to interpret them as isolation guarantees.

\section{Additional Experimental Analyses}
\subsection{Validation-based commitment and held-out evaluation}
\label{sec:valnoise}

On DS-1000 the select-best client scores \num{69}\% on validation
and \num{59.5}\% on test, whereas EvolveNet's merged harness scores
\num{63}\% on validation and \num{68.5}\% on test. The validation
and held-out scores rank these realized programs differently.
EvolveNet uses validation to decide shared updates and a disjoint
test set to evaluate the final harness.

\subsection{Comparison with centralized evolution}
\label{sec:central}

We compare against two centralized budgets on BIRD. At \emph{equal serial depth} ($T{=}3$ optimizer rounds, one proposer per round), a per-round-gated optimizer over the pooled \num{100}-item workload reaches \num{68.7}\% held-out accuracy. Chaining generations without per-round selection produces a final candidate below the unevolved harness on its training criterion and is rolled back. At \emph{equal total proposer budget}, a centralized population uses five parallel branches per round, matching EvolveNet's fifteen proposer sessions, and reaches \num{67.3}\%. Its final selected program does not improve on its first-round best branch. EvolveNet reaches \num{70.7}\%, with win \num{14} / lose \num{9} relative to the population baseline on paired items.

Inspection of the recorded artifacts shows that the centralized optimizer's strongest mechanism helps one database family while misleading others, whereas the collaborating clients develop five specialized mechanism families concurrently.

The population comparison matches proposer counts while changing
workload partitioning, specialist instructions, and aggregation.
It evaluates pooled search against workload-specialized search
followed by composition, with aggregation adding computational cost.

\subsection{A second model stack}
\label{sec:modelablation}

We repeat the BIRD pipeline with identical shards, specialist instruction, merge rule, gate, and budget, replacing the solver, proposer, and merger with MiMo-V2.5. The round-1 merge is accepted, and a later harmful merge is rejected and rolled back. The final harness reaches \num{72.7}\%, compared with \num{72.0}\% for select-best and \num{62.7}\% for the unevolved harness. Against the unevolved harness, the paired counts are win \num{23} / lose \num{8}; against select-best, the margin is \num{+0.7} points with win \num{10} / lose \num{9}. The second stack therefore also exhibits accepted gains, rejected regressions, and a final score above both baselines. The near-tie with select-best should be distinguished from the larger improvement over the unevolved harness.

\subsection{Parallel trajectories and computational cost}
\label{sec:cost}

A round runs $K$ local-evolution sessions in parallel, followed by aggregation and validation, with at most one merge revision. For the local phase, the idealized ratio between serialized and concurrent execution of the same sessions is $\sum_k C_k/\max_k C_k\in[1,K]$. In the reported runs, serialization would take \num{1.93}$\times$ ($K{=}2$), \num{3.96}$\times$ ($K{=}5$), and \num{5.37}$\times$ ($K{=}7$) as long as the measured parallel local phase. The gap from $K$ reflects uneven local runtimes; aggregate resource costs remain.

End to end, a three-round BIRD run takes \num{6110}\,s at $K{=}5$, compared with \num{1693}\,s for the per-round centralized optimizer and \num{3655}\,s for the equal-budget centralized population. These runs trade longer end-to-end runtime for higher held-out performance. Parallelism reduces local-search latency, while composition adds work. The corresponding held-out scores are \num{70.7}\%, \num{68.7}\%, and \num{67.3}\% (Sec.~\ref{sec:central}). Gate measurements reuse cached solver calls and add seconds in these runs.

Each round has one scheduled aggregation phase, whose input grows
with the number of client deltas and their execution evidence.
The transmitted programs contain a few hundred lines in these runs;
communication also includes full execution traces and verdicts.
Figure~\ref{fig:kscaling} compares the tested partitions of the fixed
workload. At $K{=}7$, the mixed shard is split into database-specific
clients of 6--8 items. Changing $K$ changes both workload granularity
and the total number of proposer sessions.

\subsection{Qualitative composition examples}
\label{sec:qualitative}

The merge reports make the operator concrete. On BIRD, the server received a value-grounding probe hard-coded to one client's database, generalized it into a dynamic categorical-value probe verified on three databases, and adopted another client's SQL-dialect auto-fix into a simpler retry structure. It rejected four shard-specific prompt rules with reference to the graded questions they broke. On DS-1000, a round-2 merge conditioned one client's schema pre-analysis on Pandas while promoting execution-retry logic with cross-library support to shared behavior. The reports cite graded verdicts or traces to explain these adoption decisions.

\clearpage
\section{Detailed Result Tables and Trajectories}
Table~\ref{tab:bylib} gives the library-level breakdown. The client-count comparison and cross-round validation trajectories are shown in the main text (Figures~\ref{fig:kscaling} and~\ref{fig:rounds}).
\begin{table*}[htbp]
\centering
\caption{DS-1000 held-out accuracy by library (\%).
Delegation, GLOBAL-only, and EvolveNet share a final-round client
snapshot; select-best is an end-to-end baseline.
$n$ denotes the number of test items.}
\label{tab:bylib}
\small
\setlength{\tabcolsep}{5pt}
\begin{tabular}{lrccccc}
\toprule
Library & $n$ & Unevolved & Select-best & Delegation & GLOBAL-only & \textbf{EvolveNet} \\
\midrule
Pandas      & 63 & \num{57.1} & \num{65.1} & \num{74.6} & \num{73.0} & \textbf{\num{73.0}} \\
Numpy       & 47 & \num{51.1} & \num{55.3} & \num{55.3} & \num{61.7} & \textbf{\num{61.7}} \\
Matplotlib  & 30 & \num{76.7} & \num{63.3} & \num{60.0} & \num{80.0} & \textbf{\num{83.3}} \\
Sklearn     & 24 & \num{45.8} & \num{54.2} & \num{62.5} & \num{45.8} & \num{58.3} \\
Scipy       & 21 & \num{42.9} & \num{57.1} & \num{57.1} & \num{61.9} & \textbf{\num{61.9}} \\
\midrule
\multicolumn{7}{l}{\emph{no client trained on this library}} \\
Pytorch     & 15 & \num{53.3} & \num{53.3} & \num{66.7} & \num{66.7} & \textbf{\num{66.7}} \\
\midrule
All         & 200 & \num{55.5} & \num{59.5} & \num{64.0} & \num{66.5} & \textbf{\num{68.5}} \\
\bottomrule
\end{tabular}
\end{table*}

\clearpage
\section{Model Access and Asset Sources}
\label{app:assets}

\subsection{Model access and execution environment}
\label{app:model-apis}

The solver, local proposer, and merger use API-accessed frozen LLMs:
DeepSeek V4 Flash for the main experiments and MiMo-V2.5 for the
second model stack. Table~\ref{tab:model-apis} lists the providers'
API entry points. Provider documentation is available from
\href{https://api-docs.deepseek.com/}{DeepSeek} and
\href{https://platform.xiaomimimo.com/}{Xiaomi MiMo}.
No model training or weight updates are performed.

\begin{table}[h]
\centering\small
\caption{Provider API entry points for the evaluated model stacks.}
\label{tab:model-apis}
\vspace{5pt}
\begin{tabular}{@{}lll@{}}
\toprule
Model & Provider & API base URL \\
\midrule
DeepSeek V4 Flash & DeepSeek & \url{https://api.deepseek.com} \\
MiMo-V2.5 & Xiaomi & \url{https://api.xiaomimimo.com/v1} \\
\bottomrule
\end{tabular}
\end{table}

Harness orchestration and task evaluation use Python processes on
Linux. DS-1000 uses a separate Conda environment with its benchmark
scientific libraries, and SWE-bench evaluates patches in
benchmark-specific Docker images. Local proposals and server merges
are generated through a non-interactive Claude Code wrapper
configured with the selected model. API credentials are supplied
through environment variables rather than included in the manuscript.

\subsection{Asset attribution and licenses}
\label{app:dataset-licenses}

The benchmark and method papers are cited in
Sections~\ref{sec:method} and~\ref{sec:protocol}.
ClawEval is credited to
\href{https://arxiv.org/abs/2604.06132}{Ye et al. (2026),
\emph{Claw-Eval: Toward Trustworthy Evaluation of Autonomous Agents}},
and the OpenClaw runtime to the
\href{https://github.com/openclaw/openclaw}{OpenClaw contributors}.
The model services are provided by DeepSeek and Xiaomi, with source
links in Appendix~\ref{app:model-apis}.

Table~\ref{tab:dataset-licenses} summarizes the licenses and terms
stated by the upstream projects. The BIRD source is the official
development set. Third-party task content remains subject to its
original license or source-platform terms; benchmark-code licenses
do not override those terms. API access is subject to the respective
providers' service terms, and no model weights are redistributed.

\begin{table}[h]
\centering\small
\renewcommand{\arraystretch}{1.15}
\setlength{\tabcolsep}{5pt}
\caption{Benchmark and runtime licenses and source terms.}
\label{tab:dataset-licenses}
\vspace{5pt}
\begin{tabular}{@{}p{0.22\linewidth}p{0.49\linewidth}p{0.23\linewidth}@{}}
\toprule
Asset & License / terms & Source \\
\midrule
BIRD development set & CC BY-SA 4.0 &
\href{https://bird-bench.github.io/}{Official project} \\
\addlinespace[2pt]
DS-1000 & CC BY-SA 4.0 &
\href{https://huggingface.co/datasets/xlangai/DS-1000}{Dataset card} \\
\addlinespace[2pt]
SWE-bench Verified & Source-repository terms for task content;
MIT for benchmark code &
\href{https://huggingface.co/datasets/princeton-nlp/SWE-bench_Verified}{Dataset card};
\href{https://github.com/SWE-bench/SWE-bench/blob/main/LICENSE}{code license} \\
\addlinespace[2pt]
LiveCodeBench & Creative Commons (\texttt{cc} in the dataset card;
specific variant not stated) &
\href{https://huggingface.co/datasets/livecodebench/code_generation_lite}{Dataset card} \\
\addlinespace[2pt]
ClawEval & MIT &
\href{https://huggingface.co/datasets/claw-eval/Claw-Eval}{Dataset card} \\
\addlinespace[2pt]
OpenClaw & MIT &
\href{https://github.com/openclaw/openclaw/blob/main/LICENSE}{Repository license} \\
\bottomrule
\end{tabular}
\end{table}

\clearpage
\section{Scope Isolation: Conditions and Proofs}
\label{app:scope_isolation}

Fix a broadcast base, mechanism implementations, and an integration
rule. Let $h(S,\tau)$ denote the executable program obtained by
incorporating mechanism set $S$ under scope assignment $\tau$.

Let $\kappa(x)$ be an observable dispatch key whose computation is unchanged by the adopted mechanisms. For the next two propositions, composition is assumed to preserve the stated guards and all pre-existing behavior outside the guarded additions. The same initial environment and realization $\omega$ are used when comparing executions.

\begin{proposition}[Scope isolation]
\label{prop:isolation}
Let $m$ be adopted with scope $\textsc{home}(d)$. Suppose that every added computation and side effect of $m$, including initialization and changes to prompt construction, is confined to a guard testing $\kappa(x)=d$; that it changes no state subsequently read outside that guard; and that adding it does not alter code or dispatch outside the guard. Then, for every $x$ with $\kappa(x)\ne d$,
\[
h(S\cup\{m\},\tau)(x;\omega)=h(S,\tau)(x;\omega).
\]
\end{proposition}
\begin{proof}
For such an input the guard is false. All computations contributed by $m$ are therefore unreachable, and the assumptions exclude changes to any code, prompt, or state observed on the remaining execution. With the same initial state and coupled randomness, the two executions follow the same trace and return the same answer.
\end{proof}

\begin{proposition}[Conflicts under disjoint scopes]
\label{prop:conflict}
Let $m_i$ and $m_j$ have scopes $\textsc{home}(d_i)$ and $\textsc{home}(d_j)$ with $d_i\ne d_j$. Suppose each satisfies Proposition~\ref{prop:isolation} both before and after the other is incorporated. For every input with $\kappa(x)=d_i$, the joint composition agrees with the composition containing only $m_i$; for $\kappa(x)=d_j$, it agrees with the composition containing only $m_j$, under the same execution realization.
\end{proposition}
\begin{proof}
On an input with dispatch key $d_i$, the addition of $m_j$ is off-domain. Apply Proposition~\ref{prop:isolation} with $S\cup\{m_i\}$ as the base mechanism set. The other case is symmetric.
\end{proof}

The propositions characterize when domain conditioning prevents off-domain interference, even if the adaptations change the same behavior within their respective domains. They do not establish that an LLM-generated composition satisfies those conditions, nor that a home-scoped edit improves performance at home. Shared-state interactions, altered prompt text, or changes to dispatch can violate the premises. Behavioral validation can expose their consequences on evaluated items, but passing a finite gate does not certify isolation for all executions. Conflicts within one scope and conflicts involving shared behavior still require an explicit composition decision.

\clearpage
\section{Proofs and Scope of the Interference Analysis}
\label{app:theory}

\subsection{Formal objects and sufficient program conditions}

The analysis fixes a finite mechanism index set $\mathcal M$, one probability law for $(Z,\Omega)$, and measurable score functions $q_A\in[0,1]$ for all $A\subseteq\mathcal M$. The input $X$ is the observable part of $Z$. Randomness is coupled to compare variants, but each marginal must remain the program's actual execution law; no independence across variants is assumed. Every evaluation begins from the specified initial environment, so persistent cross-evaluation side effects must either be reset or included in that law. The results also hold for deterministic empirical averages on a fixed evaluation set, but an empirical instantiation is not a population guarantee.

A composition family fixes both source implementations and a rule for integrating them. The order of integration is fixed even if edits do not commute. Removing a block has a specified fallback; edits that require one another to execute are bundled into one block. Therefore the mathematical set argument does not assume that arbitrary raw diff hunks are independently executable. Changing an activation scope or rewriting a block creates a different composition family. A theorem about one family cannot silently be applied to original client programs with different behavior.

For each block $i$, its effective activation region $Q_i$ satisfies, simultaneously for every background $A$ not containing $i$,
\begin{equation}
 X\notin Q_i\quad\Longrightarrow\quad
 q_{A\cup\{i\}}(Z,\Omega)=q_A(Z,\Omega)
 \quad\text{almost surely}.
 \label{eq:theory_inert}
\end{equation}
A sufficient source-level construction leaves all existing dispatch, initialization, and code outside a fixed input guard unchanged, and confines \emph{all} effects of adding the block to that guard, uniformly in the other enabled blocks. This includes prompts, imports, shared state, random-number consumption, exceptions, and budget effects. The guard depends on an unchanged observable input key. Under these conditions, off-guard traces coincide by the same argument as Proposition~\ref{prop:isolation}. An inactive guard in one observed trace is not enough to establish Eq.~\ref{eq:theory_inert}; the condition concerns the whole family. GLOBAL blocks can take $Q_i$ to be the entire input space, which is valid but may give an uninformative worst-case bound.

The quantities $g_i(A)$ and $\beta_{ij}$ are analytical quantities, not extra measurements claimed for the experiments. One always valid choice is
\[
 \beta_{ij}=\max_{A\subseteq\mathcal M\setminus\{i,j\}}
                   [-I_{ij}(A)]_+,
\]
but evaluating this expression can be expensive. Lemma~\ref{lem:theory_support} supplies a structural upper bound instead. A tighter bound requires additional semantic or evaluation evidence. No result assumes that an LLM prompt alone certifies the required properties.

\subsection{Proof of the scope-overlap bound}

\begin{proof}[Proof of Lemma~\ref{lem:theory_support}]
For fixed $A,i,j$, define the pointwise mixed difference
\[
 d_{ij}(A)=q_{A\cup\{i,j\}}-q_{A\cup\{i\}}
             -q_{A\cup\{j\}}+q_A.
\]
If $X\notin Q_i$, Eq.~\ref{eq:theory_inert} applied with backgrounds $A$ and $A\cup\{j\}$ gives $d_{ij}(A)=0$. The same holds outside $Q_j$. On $Q_i\cap Q_j$, four scores in $[0,1]$ imply $|d_{ij}(A)|\le2$. Thus
\[
 |d_{ij}(A)|\le2\mathbf 1\{X\in Q_i\cap Q_j\}.
\]
Taking expectations and using $I_{ij}(A)=\mathbb E[d_{ij}(A)]$ proves the claim. The argument is uniform in $A$. The factor two cannot generally be replaced by one: at a point with $q_A=q_{Aij}=0$ and $q_{Ai}=q_{Aj}=1$, the mixed difference is $-2$.
\end{proof}

Disjoint variable names or independent code modules alone do not prove zero score interaction: their outputs may be combined by a nonlinear scoring function. The lemma uses score-inertness outside effective activation regions, not an unjustified inference from syntactic separation to additivity.

\subsection{Proof of the composition and retention bounds}

\begin{proof}[Proof of Eq.~\ref{eq:theory_gain}]
Enumerate $S=\{i_1,\ldots,i_r\}$ and let $S_\ell=\{i_1,\ldots,i_\ell\}$, with $S_0=\varnothing$. Telescoping gives
\[
 J(C\cup S)-J(C)=\sum_{\ell=1}^r g_{i_\ell}(C\cup S_{\ell-1}).
\]
For each $\ell$, telescope the effect of previously added blocks once more:
\[
 g_{i_\ell}(C\cup S_{\ell-1})
 =g_{i_\ell}(C)+\sum_{u=1}^{\ell-1}
 I_{i_\ell i_u}(C\cup S_{u-1})
 \ge g_{i_\ell}(C)-\sum_{u=1}^{\ell-1}\beta_{i_\ell i_u}.
\]
Summing over $\ell$ counts each unordered pair in $S$ exactly once and proves Eq.~\ref{eq:theory_gain}. The background in each interaction excludes both of its indexed blocks, as required. This identity includes all dependence on previously enabled blocks, rather than truncating a set-function expansion at order two.
\end{proof}

A sometimes looser form uses only singleton gains $g_i(\varnothing)$. Telescoping through the blocks of $C$ yields
\begin{equation}
 J(C\cup S)-J(C)\ge
 \sum_{i\in S}g_i(\varnothing)
 -\sum_{i\in S}\sum_{j\in C}\beta_{ij}
 -\sum_{\substack{i<j\\i,j\in S}}\beta_{ij}.
 \label{eq:theory_singletons}
\end{equation}
This form makes cross-block interference explicit but can miss purely synergistic gains. The contextual form in the main theorem is preferable when such gains matter.

\begin{proof}[Proof of Eq.~\ref{eq:theory_retention}]
Order $O=\{i_1,\ldots,i_r\}$ and $N=\{j_1,\ldots,j_s\}$, and denote their prefixes by $O_u$ and $N_v$. The left-hand side equals
\begin{align*}
 &\sum_{u=1}^r\big[g_{i_u}(O_{u-1}\cup N)-g_{i_u}(O_{u-1})\big]\\
 &=\sum_{u=1}^r\sum_{v=1}^s
       I_{i_u j_v}(O_{u-1}\cup N_{v-1})\\
 &\ge-\sum_{i\in O}\sum_{j\in N}\beta_{ij}.
\end{align*}
The background in each summand excludes $i_u$ and $j_v$, so the contextual lower bound applies. This proves the marginal-capability retention inequality.
\end{proof}

\paragraph{Selection comparison.}
For each $C\in\mathcal C$, apply Eq.~\ref{eq:theory_gain} with new blocks $U\setminus C$. It gives $J(U)\ge J(C)+L(U\setminus C\mid C)$. Maximizing these valid lower bounds over $C$ and subtracting the actual $J_\star$ proves Eq.~\ref{eq:theory_compare}. In particular, if a maximizing candidate $C_\star$ satisfies $L(U\setminus C_\star\mid C_\star)>0$, composition strictly improves over every whole-program candidate in this family. Taking $\mathcal C$ to include the base is permitted. This is not a comparison against item-wise oracle routing, nor a claim that all mixtures or ensembles are represented by candidate subsets.

\subsection{Shared-background accumulation: proof and interpretation}
\label{app:global_home}

Corollary~\ref{cor:global_home} specializes the fixed composition family to shared blocks $C$ and disjoint home groups $S_d$. The groups may contain interacting blocks; prerequisites that must always appear together remain bundled. For $i\in S_d$, the effective-activation condition is Eq.~\ref{eq:theory_inert} with $Q_i\subseteq R_d$, uniformly over every admissible background. In particular, other home groups cannot change dispatch or introduce side effects that violate this condition. The domain regions need not exhaust the target distribution, and their masses $\alpha_d$ are determined by that distribution, not by the client training proportions. Terms with $\alpha_d=0$ are omitted, avoiding conditioning on a null event.

\begin{proof}[Proof of Corollary~\ref{cor:global_home}]
Fix $d$ and consider $X\in R_d$. For each $e\ne d$, disjointness gives $X\notin R_e$. Remove the blocks of $S_e$ one at a time from $C\cup S$. Each removal preserves the score by Eq.~\ref{eq:theory_inert} applied to the remaining background. Continuing through all $e\ne d$ yields $q_{C\cup S}=q_{C\cup S_d}$ almost surely on $R_d$. If $X$ lies outside all $R_d$, removing all home blocks similarly gives $q_{C\cup S}=q_C$. The families and domain groups are finite, so these almost-sure identities hold simultaneously after excluding a common null set. Consequently,
\begin{equation}
 q_{C\cup S}-q_C
 =\sum_{d=1}^D\mathbf1\{X\in R_d\}(q_{C\cup S_d}-q_C)
 \quad\text{almost surely}.
 \label{eq:global_home_pointwise}
\end{equation}
Taking expectations and conditioning each nonzero-mass term on its region proves Eq.~\ref{eq:global_home_accumulation}. This identity is derived from effective isolation; additivity of primitive mechanism effects is not assumed.
\end{proof}

\paragraph{Realized integration.}
Let $U=C\cup S$ and let the realized merged program $\widetilde h$ satisfy $\mathbb E|q_{\widetilde h}-q_U|\le\eta$. Then $J(\widetilde h)\ge J(U)-\eta$, and the corollary gives
\begin{equation}
 J(\widetilde h)-J(\varnothing)
 \ge [J(C)-J(\varnothing)]
      +\sum_{d:\alpha_d>0}\alpha_d\delta_d(C)-\eta.
 \label{eq:global_home_realized}
\end{equation}
This separates the integrated shared gain, conditional home gains, and realization loss. The right-hand side is positive precisely when those signed gains exceed the error bound; none is assumed positive merely because a mechanism was locally selected. For an available comparison program $c$ under the same objective, subtracting $J(c)$ instead yields
\[
 J(\widetilde h)-J(c)
 \ge J(C)-J(c)+\sum_{d:\alpha_d>0}\alpha_d\delta_d(C)-\eta.
\]
A positive right-hand side is a sufficient comparison condition, not a comparison of two unrelated lower bounds. The more general treatment of candidate mismatch follows below.

\paragraph{Shared--local interactions and scope of the conclusion.}
There is no constraint that $\delta_d(C)$ equal the gain on the broadcast base. For example, as a mathematical illustration, suppose one shared block alone and one home block alone each leave a domain score unchanged, but together improve it by $\gamma>0$. The home gain on the base is zero while its conditional gain with the shared block present is $\gamma$; Eq.~\ref{eq:global_home_accumulation} includes the resulting $\alpha_d\gamma$ contribution. The same definition allows an initially useful home block to become harmful after shared integration. Multiple blocks in $S_d$ may also interact arbitrarily. Only effects between disjoint home regions are excluded by effective isolation.

Taking $C$ as fixed does not assume that global composition is beneficial: $J(C)-J(\varnothing)$ can have either sign, and Theorem~\ref{thm:theory_composition} applies to the blocks within $C$. Nor does the decomposition prescribe a two-stage implementation; it analyzes a fixed program after its scopes and block implementations are specified. The separately generated GLOBAL-only baseline need not equal $h_C$, so its score difference from EvolveNet is not an estimate of the home sum in Eq.~\ref{eq:global_home_accumulation}. The recorded shared retry and Pandas-conditioned pre-analysis in Appendix~\ref{sec:qualitative} illustrate the distinction between $C$ and $S_d$, without certifying isolation or measuring $\delta_d(C)$ or $\eta$.

\subsection{Realized source integration and candidate mismatch}

An LLM merger may return a program $\widetilde h$ that is not exactly $h_U$. Original candidates $c_C$ may also differ from their fixed-family representatives after scopes or implementations are changed. Define nonnegative bounds satisfying
\begin{align}
 \mathbb E|q_{\widetilde h}-q_U|&\le\eta_M,\\
 \mathbb E|q_{c_C}-q_C|&\le\eta_C.
 \label{eq:theory_realization}
\end{align}
For example, each bound follows from an execution-outcome disagreement probability under the stated coupling, since scores lie in $[0,1]$. Exact realization gives zero; a small bound is not assumed to follow from a passed validation gate.

For every $C$, Eq.~\ref{eq:theory_gain} implies
\[
 J(\widetilde h)-J(c_C)
 \ge L(U\setminus C\mid C)-\eta_M-\eta_C.
\]
Consequently,
\begin{equation}
 J(\widetilde h)-\max_{C\in\mathcal C}J(c_C)
 \ge\min_{C\in\mathcal C}
       \{L(U\setminus C\mid C)-\eta_M-\eta_C\}.
 \label{eq:theory_realcompare}
\end{equation}
This robust sufficient condition is more conservative than Eq.~\ref{eq:theory_compare}, but does not identify re-scoped representatives with the original client programs. It is not evidence that the reported runs satisfy a small $\eta_M$ or $\eta_C$; establishing such values requires additional evidence about the realized programs.

\subsection{Behavioral forgetting is distinct from marginal interaction}

For a measurable evaluation region $R$ and disjoint $C,S$, define actual paired behavioral loss
\[
 \mathcal F_R(C,S)
 =\mathbb E\big[\mathbf 1\{X\in R\}(q_C-q_{C\cup S})_+\big].
\]
For binary outcomes this is the probability mass of previously correct outcomes that become incorrect in $R$, under the specified coupling. For deterministic empirical verdicts it is the corresponding fraction of evaluated items. By Eq.~\ref{eq:theory_inert}, adding all blocks in $S$ leaves scores unchanged outside $\bigcup_{i\in S}Q_i$, regardless of their order. Therefore
\begin{equation}
 \mathcal F_R(C,S)\le
 \Pr\left(X\in R\cap\bigcup_{i\in S}Q_i\right).
 \label{eq:theory_forgetting}
\end{equation}
For a realized merge differing from $h_{C\cup S}$ by expected absolute score at most $\eta_M$, the right-hand side increases by at most $\eta_M$, since the positive-part function is 1-Lipschitz. This is a locality-based upper bound, not a claim of zero within-scope forgetting. In particular, global blocks can affect all old domains.

A negative interaction need not imply forgetting: take $J(\varnothing)=0$ and $J(\{i\})=J(\{j\})=J(\{i,j\})=1$. Then $I_{ij}(\varnothing)=-1$ because the improvements are redundant, although no addition decreases the score. Conversely, even disjoint-scoped additions can harm their own scopes if their local gains are negative. Equations~\ref{eq:theory_retention} and \ref{eq:theory_forgetting} intentionally distinguish marginal usefulness from actual loss of previously successful behavior. Likewise, retaining a large fraction of this round's client gains does not establish retention of every earlier round's successes.

\subsection{A finite-proposal separation from pooled selection}

\begin{proposition}[Cross-domain regressions can block pooled search]
\label{prop:theory_pool}
Let $K\ge2$ equally weighted domains have base score $a$. Choose $g,b$ satisfying $0<b\le a$, $0<g\le1-a$, and $(K-1)b>g$. Suppose a raw candidate $r_k$ scores $a+g$ on domain $k$ and $a-b$ on every other domain. Suppose an isolated composition $h_\circ$ of the same local edits scores $a+g$ on every domain.

A pooled search makes $B$ independent proposals. Each is $h_\circ$ with probability $p\in[0,1]$ and otherwise is one of $r_1,\ldots,r_K$. It retains the highest-scoring complete candidate or the base, using exact global scores. It may archive all proposals, but it performs no additional recombination beyond its specified proposer. A local-search-and-compose procedure obtains all $K$ edits and returns $\widetilde h$ with $J(\widetilde h)\ge J(h_\circ)-\eta$. Then
\begin{equation}
 \mathbb E[J(\widetilde h)-J(h_{\mathrm{pool}})]
 \ge g(1-p)^B-\eta.
 \label{eq:theory_poolgap}
\end{equation}
With an exact population non-decreasing-score gate on $\widetilde h$, the same inequality holds for the committed local-search-and-compose output.
\end{proposition}

\begin{proof}
Every raw candidate has global score
\[
 J(r_k)=a+\frac{g-(K-1)b}{K}<a.
\]
Thus the pooled procedure returns the base unless it proposes $h_\circ$ at least once, in which case it returns a program of score $a+g$. The probability of at least one such proposal is $1-(1-p)^B$, so
\[
 \mathbb E J(h_{\mathrm{pool}})=a+g[1-(1-p)^B].
\]
The realized composition has score at least $a+g-\eta$. Subtraction gives the result. A population gate returns a score of at least $\max\{a,a+g-\eta\}$, which is no smaller. No independence between the two procedures is needed for the subtraction of expectations; independence is used only for the pooled proposal probability.
\end{proof}

The hypotheses are jointly realizable by bounded executable behavior. For instance, let each input contain its domain key, and define a reference action whose task score is $a$; a raw block switches to a higher-scoring action on its home key and a lower-scoring action otherwise. A guarded version takes its higher-scoring action only at home, falling back to the base elsewhere. Disjoint domain guards make their joint composition executable with score $a+g$ in every domain. Alternatively, these scores can be realized as success probabilities using an auxiliary uniform input variable.

The proposition is an existence result for a specified proposal process. Local discovery of all $K$ edits is a hypothesis, not a theorem about an LLM. Pooled search is explicitly allowed to produce the same guarded composite ($p>0$), so the separation is not based on forbidding conditionals in centralized programs. A centralized implementation that performs the same decomposition and recombination can reproduce the local-search-and-compose procedure. For $B=K$, the numbers of proposals match, but the latter still incurs aggregation cost. No value of $p$, $g$, $b$, or $\eta$ is estimated from the paper's reported accuracy tables. A stronger joint proposer or more pooled proposals can close the gap.

\paragraph{Forgetting versus stagnation.}
If a procedure commits raw $r_k$, each off-home domain loses $b$ relative to the base. If it instead uses the exact pooled gate, that candidate is rejected and this potential gain is not acquired. These are different consequences of the same cross-domain regression. The proposition proves the stagnation alternative for its selector; it does not claim that all centralized optimization must forget or stagnate.

\subsection{A multi-round consequence with explicit commitment error}

The preceding bounds are conditional on a round's realized mechanisms. Their multi-round use needs one fixed population objective and explicit accounting for the gate. Let $h_t$ be the committed program at round $t$ and $\bar h_t$ its final tested proposal, after any revision. Apply Theorem~\ref{thm:theory_composition} relative to $h_t$ and let $L_t$ be its lower bound for the chosen ideal composition. Suppose integration error is bounded by $\eta_t$, so $J(\bar h_t)-J(h_t)\ge L_t-\eta_t$. Suppose the actual gate chooses a maximizer of an empirical \emph{mean} score over $\{h_t,\bar h_t\}$, and, for those two programs,
\[
 |\widehat J_t(h)-J(h)|\le\epsilon_t.
\]
All quantities may depend on previous rounds and the current proposals. On any realization where these conditions hold,
\begin{equation}
 J(h_T)-J(h_0)\ge
 \sum_{t=0}^{T-1}\left([L_t-\eta_t]_+-2\epsilon_t\right).
 \label{eq:theory_rounds}
\end{equation}
Indeed, maximizing a score with uniform error $\epsilon_t$ returns a program with population score at least $\max\{J(h_t),J(\bar h_t)\}-2\epsilon_t$. Subtract $J(h_t)$, use the proposal bound, and telescope. Taking expectations is valid whenever the premises hold almost surely with integrable bounds. The appearance of $\epsilon_t$ cannot be removed merely because the validation slice is reused.

The final-tested-proposal convention matters: when the original proposal is rejected and a revision is evaluated, the displayed bound is applied to the revision, not to a potentially better rejected candidate. A general continuous-score rule based on counts of improvements and regressions need not maximize the empirical mean, so it is not covered by this corollary. This is a mathematical gate condition, not an assertion about a particular benchmark implementation.

For completeness, one sufficient statistical condition for a common $\epsilon$ is a finite class $\mathcal H_0$ fixed independently of an iid validation sample of size $n$, containing every program that can be selected. For bounded, consistently defined scores, the bounded-variable exponential inequality gives $\Pr(|\widehat J(h)-J(h)|>\epsilon)\le2e^{-2n\epsilon^2}$ for each fixed $h$. A union bound yields simultaneous validity with probability at least $1-\delta$ for
\[
 \epsilon=\sqrt{\frac{\log(2|\mathcal H_0|/\delta)}{2n}}.
\]
The simultaneous statement permits adaptive selection \emph{inside that pre-specified class}. Replacing $|\mathcal H_0|$ by the number of adaptively generated programs seen after examining validation is invalid. The class of bounded-length programs can be enormous, making this bound vacuous; we do not use it to claim a numerical certificate for the experiments.

\subsection{Counterexamples that delimit the analysis}

\paragraph{Higher-order interaction.}
For three blocks, take
\[
 J(A)=\tfrac12+\tfrac1{20}|A|-\tfrac15\mathbf1\{A=\{1,2,3\}\}.
\]
All scores lie in $[0,1]$. Every pair has zero interaction at the empty base, but $I_{12}(\{3\})=-1/5$. The three-way score is $9/20$, less than the singleton-sum prediction $13/20$. Hence empty-base pair tests cannot justify uniform $\beta_{ij}=0$.

\paragraph{Positive synergy.}
Set $J(\varnothing)=J(\{1\})=J(\{2\})=0$ and $J(\{1,2\})=1$. All contextual interactions are nonnegative, so $\beta_{12}=0$. The singleton-only bound is loose, whereas Eq.~\ref{eq:theory_gain} with anchor $C=\{1\}$ proves an increment of one.

\end{document}